\documentclass[sigconf,balance=false]{acmart}
\usepackage{popets}

\setcopyright{popets}
\copyrightyear{YYYY}
    
\acmYear{YYYY}
\acmVolume{YYYY}
\acmNumber{X}
\acmDOI{XXXXXXX.XXXXXXX}
\acmISBN{}
\acmConference{Proceedings on Privacy Enhancing Technologies}
\usepackage{multirow}
\usepackage{svg}
\usepackage{amsmath}
\usepackage[english]{babel}
\usepackage[utf8]{inputenc}
\usepackage[autostyle, english = american]{csquotes}\MakeOuterQuote{"}
\usepackage{booktabs}
\usepackage{etoolbox}
\usepackage{array}
\usepackage{tabularx}
\usepackage{placeins}
\usepackage[table]{xcolor}
\usepackage{threeparttable}
\newcolumntype{Y}{>{\raggedright\arraybackslash}X}
\usepackage{afterpage, float}
\usepackage{longtable}
\usepackage[normalem]{ulem}
\usepackage{stfloats}

\AtBeginEnvironment{table}{\preto\caption{\vspace{5pt}}}

\begin{document}

\title[Overcoming Language Barriers]{Overcoming Language Barriers:\\Multilingual Analysis of the 2023 Swiss Privacy Law's Impact}


\author{Luka Nenadic}
\orcid{0009-0004-3769-7961}
\affiliation{%
  \institution{ETH Zurich}
  \city{Zurich}
  \country{Switzerland}}
\email{lnenadic@ethz.ch}

\author{David Rodriguez}
\orcid{0000-0002-0911-4608}
\affiliation{%
  \institution{Information Processing and Telecommunications Center\\ Universidad Politécnica de Madrid}
  \city{ETSI Telecomunicación}
  \country{Spain}}
\email{david.rtorrado@upm.es}

\author{Joseph A. Calandrino}
\orcid{0009-0002-7307-5014}
\affiliation{%
  \institution{Carnegie Mellon University}
  \city{Pittsburgh}
  \state{Pennsylvania}
  \country{USA}
}
\email{jcalandr@andrew.cmu.edu}


\begin{abstract}
Policymakers enact and revise privacy laws expecting meaningful benefits for their people in practice. While scholarship has measured the real-world impact of some privacy regulations---the EU and California most notably---limited empirical evidence exists for many of the more than 140 countries that have implemented some form of privacy legislation. Switzerland, a multilingual country bordered almost entirely by EU states, is one such example. 

This paper analyzes the extent to which a 2023 alignment of Swiss privacy law with EU privacy regulation affected website privacy policies in Switzerland. To address Switzerland's unique multilingual culture, we develop an LLM-based pipeline that extracts legally relevant information as document-level labels in a single inference without requiring translation. On a benchmark of 120 expert-annotated privacy policies in German, French, Italian, and English, our pipeline achieves F$_1$~scores above 0.90 for most pairs of languages and legally relevant disclosures.

Applying this pipeline to privacy policies we collected from more than 35{,}000 Swiss- and EU-facing websites before and after the 2023 privacy law revision, we find significant increases in both mandatory and voluntary disclosures of data subject rights among Swiss privacy policies. In exploring the mechanisms driving increased disclosure rates, we discover heavy use of automated privacy policy generators and find that generated policies are associated with up to 15 percentage points higher disclosure rates. These results provide large-scale empirical evidence of how regulatory change and novel drafting technologies impact the content of privacy policies in a unique multilingual environment.

\end{abstract}


\maketitle

\section{Introduction} \label{Introduction}

By 2025, more than 140 countries had implemented data protection regulations~\cite{apacible-bernardo_data_2025}. Legislators introduce and amend these laws to meet the privacy expectations and needs of their people. Across countries, privacy laws routinely require transparent disclosure of data processing practices. Evaluating whether privacy laws achieve their intended objectives requires systematic monitoring of their real-world effects. Existing empirical research has focused heavily on the European General Data Protection Regulation (\textit{GDPR})~\cite{peukert-regulatory-2022, davis-filling-2024, frankenreiter-cost-based-2022, dabrowski-measuring-2019} as well as US state-level privacy legislation---most notably California~\cite{tran_measuring_2024, hosseini-bilingual-2024}---while empirical evidence for the privacy laws of other jurisdictions remains strikingly rare. One potential explanation for this scarcity is the substantial measurement challenges associated with systematically assessing legal compliance across jurisdictions, particularly in multilingual settings. 

Against this backdrop, this paper evaluates a major revision of Switzerland's Federal Act on Data Protection (\textit{FADP}) that entered into force in September 2023~\cite{swiss-confederation-federal-2020}. The reform aligns Swiss privacy law more closely with the GDPR but maintains differences in the specific requirements for transparency disclosures. Most notably, while Swiss privacy law offers data subjects the same rights as the GDPR---such as the right of data access---it does \textit{not} mandate that these rights be explicitly mentioned in privacy policies (\textit{policies}), contrary to the GDPR. This distinction provides a unique opportunity to examine whether websites' responses to the revision voluntarily extend beyond domestic obligations. If so, this could be interpreted as consistent with the ``Brussels Effect''~\cite{bradford-brussels-2020}, in which European Union (\textit{EU}) regulations like the GDPR exert influence beyond the EU. Overall, the Swiss privacy law revision constitutes a clearly defined regulatory intervention that allows us to examine whether changes in legal requirements are reflected in observable policy disclosures. We thus ask the following research question: \textit{Did~the 2023 Swiss privacy law revision significantly impact disclosures in Swiss website privacy policies?}

Switzerland's multilingual setting makes it both an interesting and challenging research environment. To overcome the limitations of traditional methods tailored to the English language (see Section~\ref{sec:computational_analyses}), we introduce a novel LLM-based pipeline that extracts legally relevant information as document-level labels in a single inference without requiring translation. To evaluate the pipeline's performance, we develop an expert-annotated dataset of 120 policies in German, French, Italian, and English based on an iteratively refined codebook. We annotate seven legally relevant dimensions of transparency disclosures that allow us to monitor the real-world effects of Switzerland's revision as well as potential spillover effects of the GDPR.

Finally, we inspect the role that so-called policy generators (\textit{generators}) played in potentially increasing transparency disclosures. While early forms of generators have been around since the 1970s~\cite{contreras-solving-2025, foster_when_2018}, empirical evidence remains scarce regarding (1) their prevalence (especially in non-English contexts) and (2) the extent to which they can actually improve privacy disclosures. We hypothesize that generators are primarily used by smaller firms and improve overall disclosure rates by aiding these organizations that may avoid expensive legal professionals. The most-used generators we identify are either free or cost a yearly fee of around \$100~\cite{swissanwalt-datenschutz-2023, privacybee-datenschutz-2025, datenschutzpartner_datenschutz-generator_2026}.

Our \textit{key finding} is that the Swiss revision increased disclosure rates for transparency requirements from both the revised Swiss law and the GDPR, which is consistent with (though not direct evidence of) a Brussels Effect. Based on our novel expert-annotated dataset, we find that our multilingual pipeline leveraging GPT-5 achieves F$_1$ scores above $0.90$ for the vast majority of disclosure dimensions and language pairs (German, French, Italian, and English) that we investigate. We also discover that 18\% of the analyzed local Swiss websites \textit{explicitly} use generators, and generator use is correlated with increases in disclosure rates of up to 15 percentage points (\textit{p.p.})---in contrast with prior literature that has questioned the viability of generators (see~Section~\ref{sec:automated_drafting}).

In summary, our paper makes \textit{three contributions}:

\begin{enumerate}
    \item We provide empirical evidence of the real-world effects of Switzerland's 2023 privacy law reform.
    \item We develop, validate, and publicly release a multilingual pipeline for disclosure assessment of policies, along with an expert-annotated benchmark dataset in four languages.\footnote{The codebook, dataset, and method are available in their entirety at: \href{https://doi.org/10.5281/zenodo.20794417}{\textcolor{blue}{https://doi.org/10.5281/zenodo.20512191}}.}
    \item We report on the proliferation and potential benefits of generators in a multilingual environment.
\end{enumerate}

The remainder of this paper is structured as follows: Section~\ref{sec:related work} summarizes related work. Sections~\ref{sec:method} and~\ref{sec:dataset} introduce the methods and dataset, respectively. Section~\ref{sec:results} explores the effects of the revision, and Section~\ref{sec:generators} examines the impact of generators. We discuss our key findings and limitations in Section~\ref{sec:discussion}. Section~\ref{sec:conclusion} concludes and considers future work.

\section{Related Work} \label{sec:related work}

\subsection{Automated Analyses of Policies} \label{sec:computational_analyses}

\subsubsection{Traditional Methods for Processing Policies}
Early approaches to the automated analysis of policies relied primarily on symbolic methods, processing statements through handcrafted lexicons, grammars, and ontologies~\cite{delalamo2022mapping}. Several representative systems illustrate the potential and limitations of this paradigm. PrivOnto~\cite{oltramari2018privonto} constructed a legal ontology to encode data practices such as collection and sharing, enabling structured queries over annotated corpora such as OPP-115~\cite{wilson-creation-2016}. PolicyLint~\cite{andow2019policylint} models disclosures as combinations of actors, actions, and data types to detect internal inconsistencies, and PoliCheck~\cite{andow2020policheck} extended this idea by distinguishing between first- and third-party recipients. Although these symbolic approaches offer interpretability and precision, they require substantial manual effort and are brittle under linguistic variation. Hybrid pipelines subsequently combined rule-based extraction with statistical classifiers, and supervised learning methods---particularly support vector machines and logistic regression---became widely adopted for identifying disclosures of data collection, sharing, and user choices~\cite{delalamo2022mapping,sathyendra2016optout,sathyendra2017choices}. Systems such as MAPS~\cite{zimmeck2019maps} demonstrate that traditional machine learning can scale to larger corpora. Other work explores completeness assessment, alignment of statements, or clustering of policy segments~\cite{costante2012completeness,liu2014alignment,massey2013policymining}.

Over time, neural architectures began to outperform traditional classifiers in policy analysis. Polisis~\cite{harkous-polisis-2018}, for instance, introduced an end-to-end deep learning pipeline for segmentation and classification, producing structured representations directly from raw policy text. Subsequent research applied neural models to specialized tasks, including question answering and detection of vague or ambiguous language~\cite{ravichander2019privacyqa,liu2016vagueness}. Despite improved predictive performance, these approaches remained heavily dependent on annotated corpora and predefined taxonomies, which limit their scalability across jurisdictions and languages. In particular, most available datasets and trained models focus predominantly on English-language policies, limiting their applicability in multilingual contexts.

\subsubsection{LLM-Based Policy Analysis}
Recent studies suggest that large language models (\textit{LLMs}) can substantially facilitate the automated analysis of policies by enabling task specification through natural-language instructions rather than extensive feature engineering and model tuning. Rodriguez et al.~\cite{rodriguez2024llmprivacy} evaluated GPT and LLaMA models, reporting that GPT-4 Turbo, under carefully crafted prompts, achieved F$_1$ scores above 93\% on multi-label classification tasks over benchmark corpora. While MAPS~\cite{zimmeck2019maps} and Polisis~\cite{harkous-polisis-2018} previously represented the state of the art in scalability and granularity, LLMs can reach comparable or even higher accuracy with substantially less task-specific engineering. In a similar vein, Tang et al.~\cite{tang2023policygpt} introduced PolicyGPT, a zero-shot framework that reached 87–97\% accuracy across multiple policy classification datasets, highlighting the continued importance of annotated corpora for validation. Goknil et al.~\cite{goknil2024papel} further demonstrated that performance varies markedly across prompt design strategies, underscoring the sensitivity of prompt-based approaches.

More recent work has extended LLM-based analysis toward legally grounded compliance assessments. Cory et al.~\cite{cory2025wordlevelannotationgdprtransparency} performed fine-grained annotation of GDPR transparency requirements using a two-stage pipeline with self-correction, evaluating eight LLMs over 703k app policies. Complementarily, Xie et al.~\cite{xie2025evaluating} proposed an LLM-based framework to evaluate policies against 34 clauses derived from the GDPR and US state laws, achieving average F$_1$ scores of 0.94 and enabling large-scale analysis of thousands of websites. While these studies illustrate how LLMs extend prior work toward more fine-grained and legally grounded compliance assessments at scale, they also highlight important limitations: both are restricted to English-language policies and explicitly identify multilingual evaluation as a key direction for future research.

\subsection{Datasets for Policy Analysis}

Progress in the automated analysis of policies has been closely tied to the availability of annotated datasets. For early symbolic and statistical approaches, corpora such as OPP-115~\cite{wilson-creation-2016} and APP-350~\cite{zimmeck2019maps} provided thousands of labeled statements about data collection and sharing practices. MAPP~\cite{arora-tale-2022} extended this line to multilingual contexts (English and German), while IT-100~\cite{guaman2023crossborder} targeted GDPR-specific disclosures of international transfers. These datasets served a dual role: training material for supervised classifiers as well as benchmarks for comparing different methods. Although LLMs typically no longer require task-specific training on these datasets, they still rely on them as a benchmark for validation~\cite{guha2023legalbench}.

Subsequent resources diversified both task framing and scale. PrivacyQA~\cite{ravichander2019privacyqa} and PolicyQA~\cite{ravichander2020policyqa} entail question-answering tasks, linking the policy text with expert-curated evidence or span-level reading comprehension. Moreover, APPCorp~\cite{liu2023appcorp} annotated document organization at paragraph level. Longitudinal corpora, such as the Princeton–Leuven dataset~\cite{amos-privacy-2021}, assembled nearly one million snapshots across a decade to study the evolution of policies post-GDPR. Other important longitudinal corpora include the work of Linden et al.~\cite{linden-privacy-2020} and Wagner~\cite{wagner2023privacypolicies}.

Recent efforts further align datasets with regulatory and jurisdictional contexts. GDPR-NER~\cite{darji2024gdprner} annotated European policies with entities defined in the Data Privacy Vocabulary~\cite{pandit-data-2024} ontology, while C3PA~\cite{story2024c3pa} encoded over 48,000 policy segments against the disclosure requirements of the California Consumer Privacy Act~(\textit{CCPA}). Marotta-Wurgler and Stein~\cite{marotta-wurgler-stein-2025-building} annotated multi-class labels capturing legally salient clause interdependencies. Other corpora expand linguistic and geographic coverage, including a Saudi dataset of annotated Arabic policies~\cite{mashaabi2023arabic}; the 100-Platforms corpus~\cite{palka2023annotated}, which samples major services for high-level data practices; and the longitudinal, multilingual scraper by Bernhard et al.~\cite{Bernhard2025scraper}. Together, these datasets illustrate a shift from general-purpose classification resources toward regulatory-aligned, multilingual corpora.

\subsection{Automated Drafting of Legal Documents} \label{sec:automated_drafting}

In an effort to comply with privacy laws, parties (especially resource-constrained ones) may turn to automated policy generators (\textit{generators}) to draft tailor-made policies at a fraction of the cost of lawyers. In legal literature, Betts and Jaep \cite{betts-dawn-2017} provide a historical overview of the use of generators by lawyers and their clients for various types of legal documents. They also explore the possibility of automation via machine learning. Contreras \cite{contreras-solving-2025} discusses how generative AI could address the shortcomings in scope and range of traditional generators. Barton and Rhode \cite{barton-access-2019} present lawsuits and regulatory restrictions that consumer-oriented legal services such as generators have faced in the US. Many theoretical contributions from legal scholars often discuss generators under the umbrella term of ``document assembly systems''~\cite{foster_when_2018, saxon_computer-aided_1982}.

Empirical data related to generators is scarce. To our knowledge, only two studies have illuminated some key aspects of generator use ``in the wild.'' First, Sun and Xue~\cite{sun_quality_2020} investigate the output quality of policy generators by drafting a total of thirty policies for three synthetic apps and ten generators. Second, Pan et al.~\cite{pan_is_2024} examined nearly 50,000 app policies on Google Play and found that generators have likely been used to generate 20.1\% of the policies. Similar to the first study, the authors generate synthetic policies to explore compliance issues in the generated documents.

\section{Method} \label{sec:method}

The analysis of policies poses a persistent challenge, as they are lengthy, vague, and full of legal jargon, which complicates automated processing~\cite{delalamo2022mapping, rodriguez2024sharing}. In Switzerland, like in the EU, this difficulty is compounded by the multilingual environment. As a result, any empirical assessment of disclosures must be able to accommodate this linguistic diversity without introducing biases.

Traditional computational approaches (see Section~\ref{sec:computational_analyses}) are poorly suited to this task. Symbolic methods and lexicon-based classifiers have been developed almost exclusively for English. Supervised learning approaches depend on annotated corpora that exist primarily in English (e.g., OPP-115) and cannot be directly applied to other languages without extensive re-annotation. Even deep learning models such as Polisis~\cite{harkous-polisis-2018} or MAPS~\cite{zimmeck2019maps} rely on English-only training data and fixed taxonomies, reinforcing an anglophone bias~\cite{mhaidli2023researchers} that limits their adequacy in the Swiss context.

LLMs provide a more flexible alternative: a single model can be applied across different languages without retraining, avoiding the need for language-specific resources or feature engineering. At the same time, LLMs can capture compliance-relevant disclosures---such as the purposes of data processing---in a consistent manner across languages. Importantly, their performance must still be assessed against independent human-coded data to ensure the accuracy, completeness, and legal relevance of model outputs, as LLMs are not immune to decreased performance in specific languages.

\subsection{Codebook and Annotations}\label{sec:codebook}
In line with the ``text-as-data'' approach in computational social sciences~\cite{grimmer-text-2022}, we constructed a multilingual benchmark based on a systematically developed codebook. The codebook translates statutory obligations from the GDPR and the revised Swiss privacy law (\textit{FADP}) into discrete annotation questions. We use these questions to construct the human annotations that serve as the reference benchmark for evaluation as well as to guide the automated assessment, since the codebook is supplied to the LLM in the prompt.

Since translating complex legal requirements into discrete annotations is complex, we first identified obligations that are reflected in policies and can be assessed objectively. This led us to the information duties under Art.~13 GDPR, which we distilled to the following obligations that are reasonably objective to annotate: the controller's identity (including contact details) and the purposes of processing (Art. 13(1) GDPR), as well as the rights of data subjects (access \& rectification, erasure, portability, right to lodge a complaint with a supervisory authority, and transparency when performing automated decision-making; Art.~13(2) and 15--22 GDPR).

While distilling complex legal requirements into discrete annotations is necessary to create scalable compliance proxies, we concede that our procedure cannot estimate disclosure quality. Some policies may, for example, add a user-friendly summary of the data subjects' rights to help users understand and invoke their rights.

Our annotated transparency obligations align with prior legal scholarship~\cite[see online appendix, p. 17]{frankenreiter-cost-based-2022}. We extend this work in two ways: we add the controller's identity and contact information and the purposes of data processing as additional transparency disclosures, and we exclude non-binary disclosures such as the data retention period (Art. 13(2)(a) GDPR). This ensures our questions are both (1) validated by previous work and (2) legally grounded.

While this approach yields fewer annotated questions than Cory et al.~\cite{cory2025wordlevelannotationgdprtransparency}, our questions still serve as robust and scalable proxies for estimating the effect of the 2023 Swiss privacy revision on Swiss website policies. The smaller set also offers a practical advantage: we can pass all codebook questions to the model in a single inference pass. This mitigates known performance degradation from longer prompts~\cite{levy_same_2024, xie2025evaluating} and substantially reduces costs compared to running multiple inference rounds.

The GDPR's data subject rights do not perfectly match the FADP: most notably, the revised FADP does \textit{not} oblige controllers to inform data subjects about their rights: access \& rectification, erasure, portability, and lodging a complaint. For the purposes of our annotations, these differences do not matter: any given policy either mentions a specific right (e.g., ``You have the right to delete your data.'') or not. Since these disclosures are only required under the GDPR, their explicit mention in Swiss-facing websites is consistent with, though not direct causal proof of, the Brussels Effect.

Table~\ref{tab:codebook} lists the full set of questions and coding instructions. We annotate only the controller’s practices as stated in the policy and ignore sections unrelated to Switzerland or the EU (e.g., clauses related to California's privacy law). The unit of analysis is the entire policy: each item is coded ``1'' if the policy explicitly mentions the corresponding element, ``0'' otherwise; the ``last updated'' field is coded as the date provided in the policy text or ``NA'' if absent.

We developed the codebook iteratively in three phases with three annotators: (1) a lawyer with an advanced law degree; (2) a law student nearing completion of an advanced degree; and (3) a privacy engineer (Ph.D.) with five years of work on GDPR-related topics. Edge cases and instructions were discussed among all three throughout the iterative codebook refinement procedure.

The first phase, including 15 English policies, yielded promising reliability metrics (Krippendorff’s $\alpha = 0.74$--$1.0$). Disagreements led us to refine the controller information (\texttt{contr}), data portability (\texttt{port}), and automated decision-making (\texttt{hum}) questions. A second phase with 20 policies produced $\alpha$-values between 0.67 and 0.93. Following standard guidelines~\cite{krippendorff2004reliability}, values above 0.80 indicate strong reliability, while values between 0.67 and 0.80 are generally considered acceptable.

\begin{table*}[ht]
\centering
\begin{threeparttable}
\begin{tabular}{llllc}
\toprule
\textbf{Code} & \textbf{Codebook question} & \textbf{Legal basis} & \textbf{Disclosure required}\tnote{1} & \textbf{Kripp. $\alpha$}\tnote{2} \\
\midrule
\texttt{ispol} & Is the text a privacy policy? & -- (screening) & -- (screening) & 0.672 \\
\texttt{upd}   & Date of last update & -- (metadata) & -- (metadata) & --\tnote{3} \\
\texttt{contr} & Controller's identity and contact & Art.~13(1)(a) GDPR /Art.~19(2)(a) FADP & GDPR and FADP & 0.672 \\
\texttt{purp}  & Purposes of processing & Art.~13(1)(c) GDPR /Art.~19(2)(b) FADP & GDPR and FADP & 0.825 \\
\texttt{rect}  & Right of access and rectification & Art.~15--16 GDPR / Art.~25 FADP & GDPR only & 0.839 \\
\texttt{forg}  & Right to erasure & Art.~17 GDPR / Art.~30 FADP & GDPR only & 0.929 \\
\texttt{port}  & Right to data portability & Art.~20 GDPR / Art.~28 FADP & GDPR only & 0.868 \\
\texttt{comp}  & Right to lodge a complaint & Art.~77 GDPR / Art.~49(1) FADP & GDPR only & 0.933 \\
\texttt{hum}   & Automated decision-making & Art.~22 GDPR / Art.~21 FADP & GDPR and FADP & 0.825 \\
\bottomrule
\end{tabular}
\begin{tablenotes}
\item[1] Requirements from both the GDPR and FADP are italicized in all tables throughout the paper.
\item[2] Values > 0.80 indicate strong reliability, while values between 0.67 and 0.80 are considered acceptable~\cite{krippendorff2004reliability}.
\item[3] Not applicable for metadata item.
\end{tablenotes}
\end{threeparttable}
\medskip
\caption{Codebook questions, legal bases, and inter-annotator reliability metrics (Krippendorff's $\alpha$ in second phase).}
\label{tab:codebook}
\end{table*}

\begin{figure*}[ht]
\centering
\includegraphics[width=0.75\linewidth]{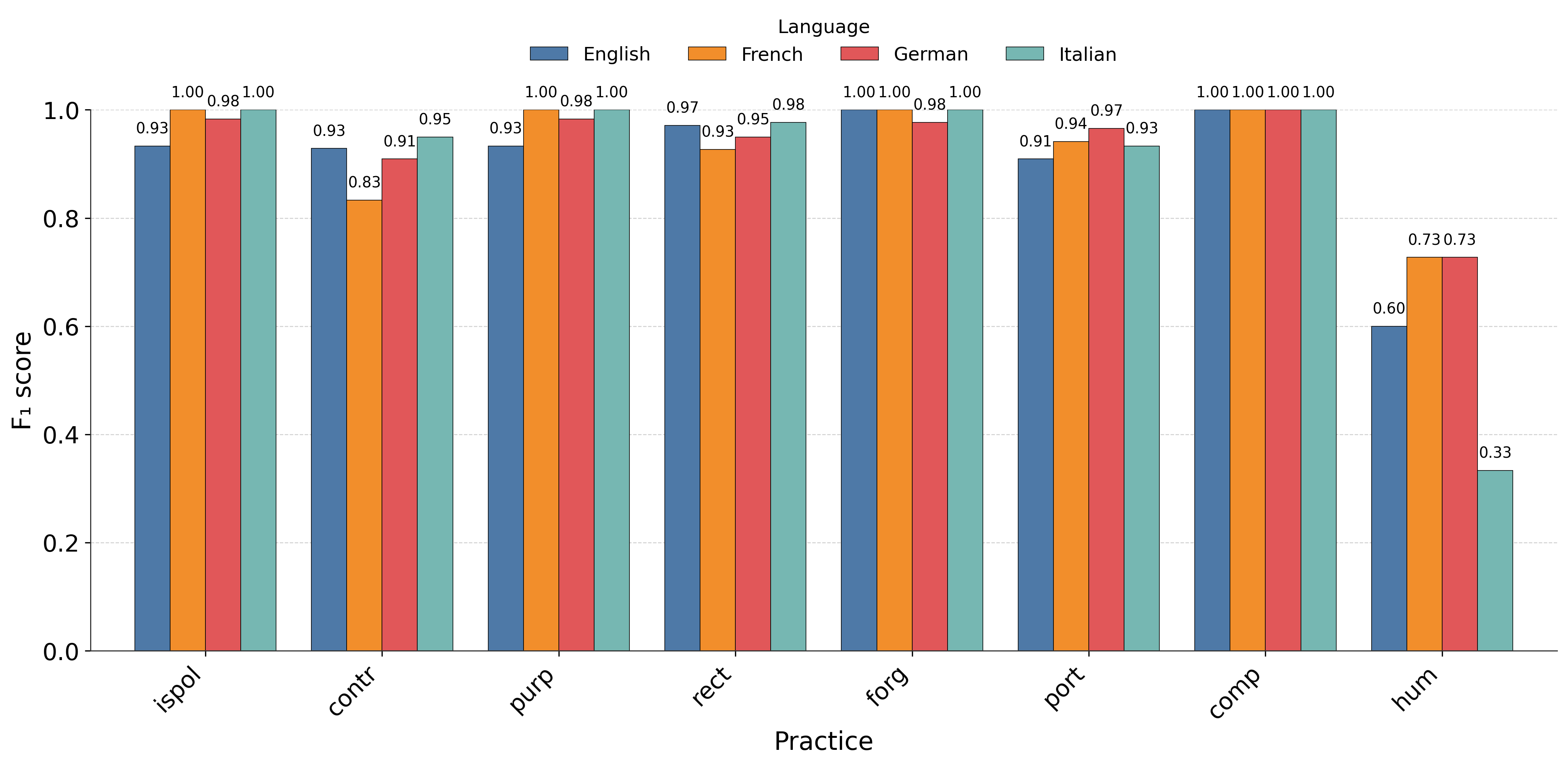}
\caption{F$_1$ score of GPT-5 by practice and language (with annotated dataset as benchmark).}
\Description{Bar chart showing F$_1$ scores of GPT-5 across English, French, German, and Italian for each disclosure practice. Most scores are above 0.9, except controller information in French and automated decision-making.}
\label{fig:f1-gpt5}
\end{figure*}

All retained dimensions reached the acceptable threshold; most exceeded 0.80 (see Table~\ref{tab:codebook}).\footnote{We dropped one codebook question (related to data protection officer) due to inconsistent terminology across policies.} Finally, the updated codebook was applied to 120 policies (30 each in German, French, Italian, and English). Each policy was independently annotated by a single coder proficient in the respective language, a common practice once high agreement has been reached~\cite{wong2021cross, artstein2008intercoder}, which we did in the second round. The 30 policies per language align with prior work~\cite{zimmeck_automated_2017}, where validation sets typically involve only a few dozen policies, as policy annotations are time-intensive.

To ensure consistency across phases, all policies used in the three annotation rounds were drawn from the customized dataset introduced in Section~\ref{sec:dataset}. In short, the final dataset consists of the German, French, Italian, and English policies of 35,000 websites collected in August and October 2023. Random sampling was performed from the August 2023 snapshot, independently for each phase. This approach formally allowed overlaps, though in practice only one repetition occurred. These subsets provided a manageable basis for iterative refinement of the codebook and for establishing a multilingual benchmark to validate the method. It also allowed us to annotate the policies closer to a real-world setting, with most transparency disclosures occurring in a balanced way (45--79 positive instances out of the 120 annotated policies), with the exceptions of the purposes of processing (100 positive instances) and automated decision-making (15 positive instances) disclosures.

Beyond serving as a benchmark for validation, our dataset itself constitutes a contribution. Established resources such as OPP-115 or APP-350 remain valuable references and have been widely used~\cite{zimmeck2019maps, harkous-polisis-2018, andow2019policylint}. However, their widespread availability raises uncertainty about whether they were incorporated into LLM pretraining, creating a risk of overfitting when used for validation. Our dataset addresses this concern and provides a novel multilingual legal benchmarking resource. We release the codebook and annotations to support reproducibility and enable further research.

\subsection{Method Design}

Our automated method applies OpenAI models---as Cory et al. reported their comparatively superior performance for detecting GDPR-compliance in English policies~\cite[p. 520]{cory2025wordlevelannotationgdprtransparency}---to classify policies according to the dimensions defined in our codebook. Throughout this section, we refer to each of the nine dimensions by the abbreviated code as defined in Table~\ref{tab:codebook}. For each policy, we issue a single inference request to the model and require it to return a JSON object conforming to a predefined schema. We rely on OpenAI’s structured output functionality~\cite{openai-structured-outputs-2024}, which enforces the presence and type of all nine fields and prevents the generation of invalid formats or hallucinated attributes. The call is structured with a fixed \textit{system message} specifying the global task and rules, followed by two \textit{user messages}: (1) the codebook, and (2)~the raw policy text. We release the prompts and the schema \textit{verbatim} with our artifacts.

The user prompt mirrors the codebook by including each question along with the same decision rules and edge-case clarifications used by annotators. To further reduce ambiguity, we also incorporate few-shot examples demonstrating positive and negative cases for each item. Prior work has shown that such examples significantly improve the reliability of LLM-output in this domain~\cite{rodriguez2024llmprivacy}, and their inclusion follows standard practice in human codebook development, where annotated examples guide consistent application of coding rules~\cite{Zlabinger2020dexa, v7labs-2022-annotation-guidelines-best-practices, tseng-stent-maida-2020-annotation-best-practices}. This alignment also allows direct comparison between manual and automated annotations.

The model processes policies in German, French, Italian, and English without translation. To reduce the risk of the model relying on superficial entity cues (e.g., company name) and to mitigate privacy risks, all identifying information is replaced with placeholders using Presidio~\cite{microsoft-presidio-nodate} prior to inference. Additional global rules are enforced: if \texttt{ispol} is false, all other booleans must also be false and the date fields set to ``NA.'' Dates are normalized to the format DD/MM/YYYY for the most recent date mentioned in the policy, or ``NA'' if none is present. Figure~\ref{fig:last_updated} in Appendix~\ref{app:last_updated} provides an overview of the last updated dates for October policies.

A key feature of our approach is that the model evaluates all disclosure dimensions in a single prompt. Rather than decomposing the task into separate binary classifiers, we treat disclosure assessment as a structured, multi-dimensional extraction problem applied to the policy as a whole. This joint design preserves contextual dependencies between obligations and mirrors the way human annotators apply the codebook to an entire document. In contrast, prior work has typically addressed individual practices in isolation~\cite{rodriguez-data-2024, yang2025dbsec, rodriguez2024llmprivacy}. Beyond efficiency, this unified extraction framework promotes coherent document-level reasoning across legally related dimensions.

We do not apply fine-tuning or task-specific training, and model behavior is determined solely by the fixed prompts and schema. The alias \texttt{gpt-5-chat-latest} resolved to \texttt{gpt-5-2025-08-07}, which was used for the validation and the large-scale analysis. In repeated validation runs during development, we did not observe instability in the structured outputs for identical inputs, suggesting that schema enforcement and fixed prompting foster stable predictions.

\subsection{Method Validation} \label{sec:method_validation}
We validated our method against the independently annotated dataset of 120 policies (30 each in German, French, Italian, and English). This dataset allowed us to directly compare model predictions against human annotations. The final benchmark of 120 policies was treated as a held-out validation set. Prompt design and schema refinement were completed prior to the full validation run, and no prompt adjustments were made based on model performance on this benchmark. This separation reduces the risk of prompt overfitting to the evaluation set.

The annotator with an advanced degree in law reviewed all disagreements between manual annotations and model outputs. This annotator has intermediate Italian proficiency and full professional or native proficiency in the other three languages. When unambiguous annotation mistakes were identified in the human-annotated data (e.g., a policy with the following text: ``Art. 22 GDPR: In cases of solely automated processing with legal effects, special rights are granted for data subjects.'' was erroneously labeled as not mentioning automated decision-making), they were corrected. The full list of corrections is provided in Appendix~\ref{app:annotations_comparison}.\footnote{A total of 21 annotation errors were identified (including one misclassified policy accounting for 7 items), representing less than 2\% of all annotations.} For transparency, Appendix~\ref{app:corrections_impact} additionally reports robust model performance on the original annotations prior to correction. In our view, the benefits of an accurate benchmark for our study as well as for future use outweigh the potential issues of post-annotation corrections.

The evaluation metrics were then computed on the corrected benchmark. To assess the stability of these estimates given the sample size per language ($N = 30$), we additionally computed 95\% bootstrap confidence intervals for precision, recall, and F$_1$ by resampling policies with replacement (1,000 resamples, fixed random seed). The resulting intervals are generally narrow across all obligations, indicating stable performance estimates. As expected, uncertainty is substantially larger for \texttt{hum}, which has very few positive cases in the human-annotated dataset (3--4 per language; see Appendix~\ref{app:bootstrap-ci}).

We evaluated four OpenAI models---GPT-4o, GPT-4.1, o4-mini, and GPT-5---using identical prompts and the same schema. Figure~\ref{fig:f1-gpt5} reports the detailed F$_1$ results for GPT-5 across languages and practices. Most dimensions reach near-perfect agreement with the human annotations: \texttt{comp} (right to lodge a complaint), \texttt{ispol} (policy detection), \texttt{purp} (purposes of processing), \texttt{forg} (erasure), \texttt{port} (portability), and \texttt{rect} (rectification) all exceed 0.9 in nearly every language. \texttt{contr} (controller information) shows more variability, performing well in English (0.93), German (0.91), and Italian (0.95) but dropping to 0.83 in French. The lowest and most variable scores arise for \texttt{hum} (automated decision-making), with values ranging from 0.73 in German and French to 0.60 in English and 0.33 in Italian. This volatility likely arises from the aforementioned very low number of positive cases in the human annotations, with even a single misclassification substantially affecting the F$_1$ scores. The low prevalence is likely due to the fact that, in contrast to the other obligations, the requirement to mention \texttt{hum} is \textit{conditional} on the website performing automated decisions. Nonetheless, we must advise caution when interpreting our results for the \texttt{hum} metric.

Results for GPT-4.1 and o4-mini lag behind GPT-5. For GPT-4.1, the largest shortfalls versus GPT-5 occur in \texttt{port} (French, $-0.24$), \texttt{contr} (German, $-0.18$), and \texttt{hum} (roughly $-0.15$ in English, French, and German). Smaller but consistent drops also appear in \texttt{rect}, \texttt{forg}, and \texttt{purp}, while improvements are rare and inconsistent. Results for o4-mini mirror this profile, with similarly large losses in \texttt{port} (French, $-0.24$) and \texttt{hum} (German, $-0.20$), together with moderate declines in \texttt{contr}. Gains are again rare, and the model suffers from slower inference speed. Both models show an apparent improvement in \texttt{hum} (Italian, around $+0.2$), but---as stated---this dimension should be interpreted with caution. Taken together, these patterns make GPT-5 the more reliable and efficient option for evaluating multilingual disclosures of transparency obligations.\footnote{We provide the performance metrics of GPT-4.1 and o4-mini in our artifact repository.}

\begin{figure}[ht]
\centering
\includegraphics[width=1\linewidth]{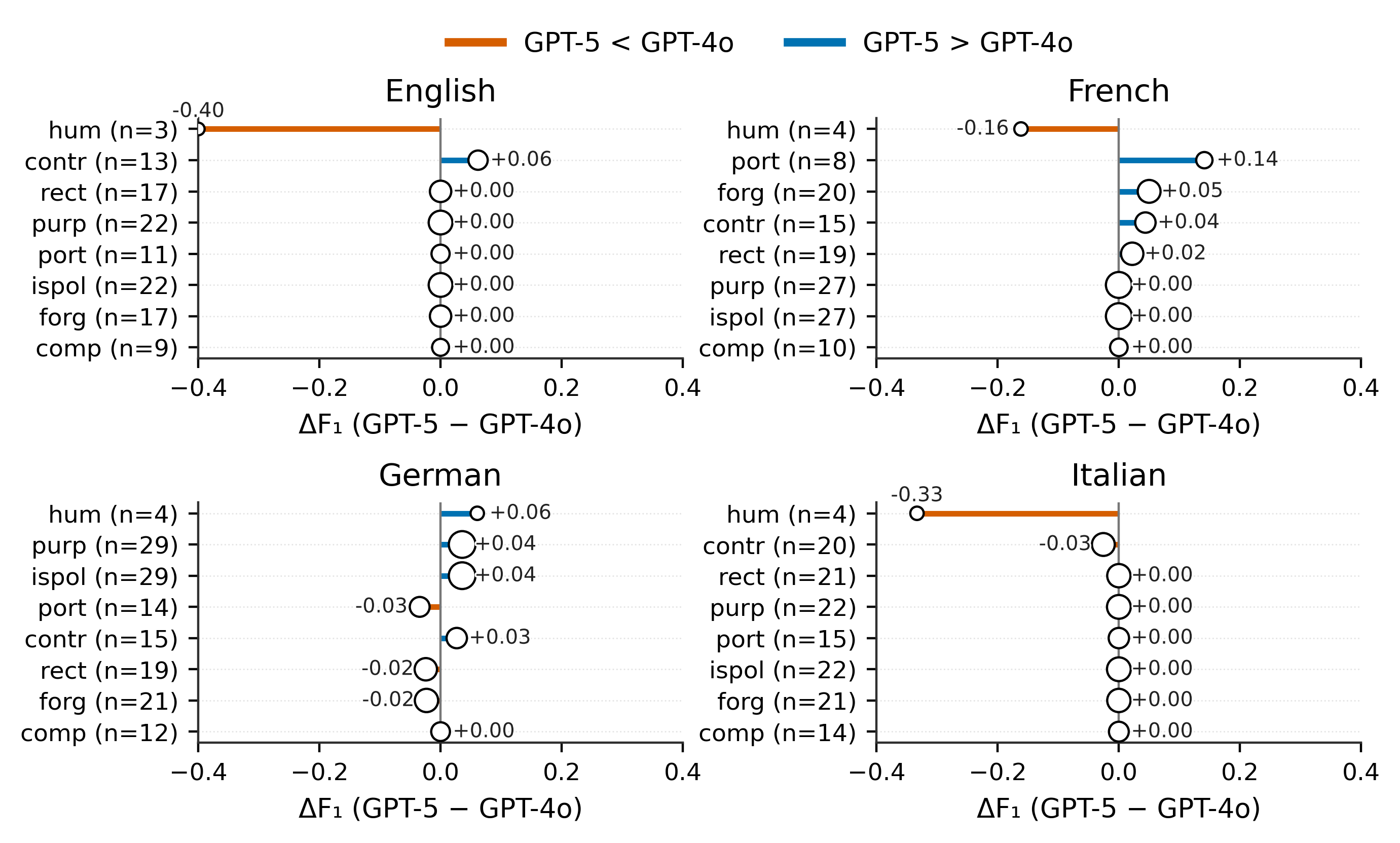}

\footnotesize{Bubble size is proportional to the number of positive human-annotated cases per practice. Labels show $\Delta$F$_1$ values. Blue bars indicate GPT-5 > GPT-4o, orange GPT-5 < GPT-4o.}
\caption{Performance difference ($\Delta$F$_1$) between GPT-5 and GPT-4o across languages and obligations.}
\Description{Cleveland dot plot showing the relative F$_1$ score difference between GPT-5 and GPT-4o. Most values are near zero, with small gains in some practices and drops in automated decision-making. Bubble size indicates the number of positive cases.}
\label{fig:delta-f1}
\end{figure}

We, therefore, narrowed down our comparison to GPT-5 and \mbox{GPT-4o}, as GPT-4o showed the closest overall performance to \mbox{GPT-5}. Figure~\ref{fig:delta-f1} visualizes their relative performance. The two models are broadly aligned, with most practices showing $\Delta$F$_1$ scores near zero. GPT-5 offers modest gains---for instance \texttt{contr} in English (+0.06), \texttt{port} in French (+0.14)---while its largest setbacks appear in \texttt{hum}, with $-0.40$ in English and $-0.33$ in Italian. Bubble size in the plot reflects the number of positive cases in the human annotations, making clear that the steepest drops occur in the practice with the lowest support. We ultimately selected GPT-5 because it offers a more favorable cost profile and ensures longer-term availability.

\section{Dataset} \label{sec:dataset}

\subsection{Initial Dataset}

The dataset for this project was created using Bernhard et al.'s~\cite{Bernhard2025scraper} scraper. The scraper automatically collects the policies and terms of service of 800,000 websites monthly in 37 languages. We refer to Table~\ref{tab:classification-metrics} in Appendix~\ref{app:scraper}, drawn and adapted from Bernhard et al.~\cite[p. 59]{Bernhard2025scraper}, for the performance of the scraper at collecting website policies in German, French, Italian, and English. Most notably, the scraper reaches recall scores ranging from 0.78 to 0.91. We also inherited the scraper's functionality of only extracting the single most likely policy per website as a target URL~\cite[Sections 3.1 and 3.4]{Bernhard2025scraper}. This approach substantially reduced false positives for the vast majority of websites, which is paramount to our analysis. The trade-off, however, is that we do \textit{not} collect multilingual or multijurisdictional versions of a website's policy if such exist.

For the purposes of this paper, we first scraped a sample of 90,000 websites stratified by popularity each in (1) Switzerland and (2) all EU countries, Iceland, Liechtenstein, Norway, and Great Britain. We collected a representative sample of websites with various levels of popularity in August 2023 according to the Chrome User Experience Report (\textit{CrUX}~\cite{chrome-overview-2024}). Due to overlaps between the Swiss group and the non-Swiss group, the procedure resulted in the initial selection of 147,029 websites, of which the scraper could access 144,690 in August and 145,193 in October. For Switzerland, we selected all top 50k websites as well as 20k randomly sampled ones from the 100k-500k and 500k-1M CrUX popularity buckets, respectively. For the non-Swiss group, we collected 500 randomly sampled websites for each popularity bucket (top 1k through 1M-5M, if available) in each country. We control for website popularity rank in our statistical analyses. The initial website budgeting procedure as well as the corresponding CrUX list can be found in our artifact repository.

Each website was scraped once one week before the entry into force of the revision of Swiss privacy law (August 23-28, 2023) and once one month thereafter (October 4-10, 2023). This relatively short time frame increases the likelihood that observed changes occurred due to the revision itself, and not as a result of subsequent events like court cases. While we acknowledge that longer-term adaptations of websites to the revision may occur more gradually, our conservative approach is consistent with Linden et al.'s~\cite[p. 50]{linden-privacy-2020} observation that policy updates clearly clustered within the first month after the GDPR's entry into force.

\subsection{Relevant Subset \& Website Categories}

We utilized only a subset of the initial dataset in our subsequent analyses. To assess whether observed changes are plausibly associated with the entry into force of the revised FADP, we compare disclosure rates across three predefined website groups. Similarly to Peukert et al.~\cite[p.~752]{peukert-regulatory-2022}, group assignment is based on (i) top-level domain (\textit{TLD}), (ii) detected policy language using CLD3~\cite{google-compact-2022}, and (iii) country-specific traffic popularity buckets (1k, 5k, 10k, 50k, 100k, 500k, 1M, and 5M) from CrUX. We restrict the analysis to policies in German, French, Italian, or English. German, French, and Italian are the official languages of Switzerland,\footnote{We do not consider the language Romansh due to its limited availability.} and English is commonly spoken. Websites are categorized as follows:

\begin{itemize}
    \item \texttt{CH}: websites with a \texttt{.ch} or \texttt{.swiss} TLD that do not appear in any EU-country popularity bucket.
    \item \texttt{CH \& EU}: websites with a Swiss TLD that also appear in at least one EU-country popularity bucket.
    \item \texttt{EU}: websites with German, Austrian, French, or Italian TLDs that do not appear in any Swiss popularity bucket.
\end{itemize}

We refer to the \texttt{CH} and \texttt{CH \& EU} groups jointly as Swiss-facing websites. Websites with generic TLDs (e.g., \texttt{.com}) are excluded due to the inability to infer geographic targeting.

Under this design, the \texttt{EU} group serves as a quasi-control group, as these websites are not directly subject to the Swiss reform. While indirect spillovers cannot be excluded, any reform-specific disclosure changes should be more pronounced in the Swiss-facing groups than in the \texttt{EU} group. We decided to only include websites from Switzerland's neighboring countries in the quasi-control group, because these websites shared a stable privacy environment (GDPR applicable since 2018) and cultural/linguistic ties with Swiss websites, whereas other websites may have introduced noise in our identification method (see Section~\ref{sec:disclosure_practices}) from unrelated events like the California privacy law revision that entered into force in 2023.

Finally, we restrict what we define as valid policies in two ways in order to minimize the occurrence of false positive documents that would undermine our results. We first discard texts that (1) have been classified as a policy with a probability of less than 50\% by the classifier introduced by Bernhard et al.~\cite{Bernhard2025scraper} or (2) contain less than 200 characters. Our second filtering consists of keeping only the texts that have been classified as a policy according to GPT-5 (see Section~\ref{sec:method}). In total, GPT-5 discarded 1,013 policy observations (485 in August and 528 in October), which corresponds to 3.6\% of the 27,804 policy observations that persisted from the first filtering step. This yields 26,791 valid policy observations across both snapshots.

\begin{figure}[t]
    \centering
    \includegraphics[width=\columnwidth]{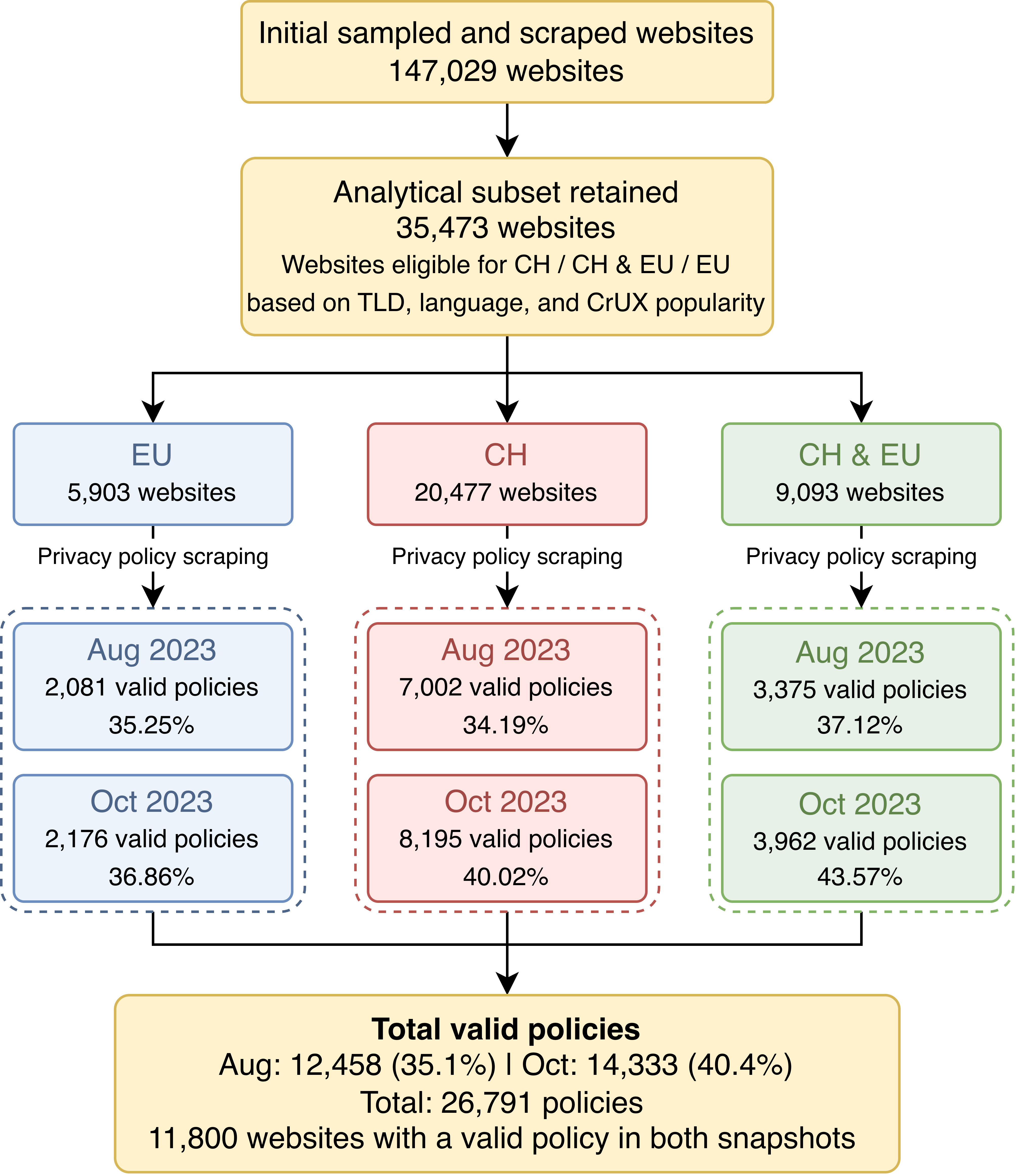}
    \caption{Construction of the analytical dataset.}
    \Description{Construction flow of the analytical dataset. We start with an initial sample of 147,029 websites, which we then group into 35,473 websites in total for the three groups CH, EU, and CH \& EU. These websites have 12,458 valid policies in August and 14,333 valid policies in October 2023.}
    \label{fig:dataset-construction}
\end{figure}

Figure~\ref{fig:dataset-construction} summarizes the entire construction of the analytical dataset and distinguishes the retained website sample from the valid policy observations used in the subsequent analyses. Each group (\texttt{CH}, \texttt{CH \& EU}, and \texttt{EU}) contains more than 5,000 websites, with German as the predominant language across all groups, particularly within \texttt{CH} (see Table~\ref{tab:languages} in Appendix~\ref{app:languages} for the full language distribution).

\section{Policies Before and After the Revision} \label{sec:results}

\subsection{Descriptive Groups Statistics} \label{sec:statistics}

Our analysis ultimately builds on a total of 35,473 websites observed at two time points: August 2023 (immediately before the entry into force of the revised FADP) and October 2023 (one month thereafter). The set of websites is held constant across both snapshots; however, the presence of a detectable policy varies between August and October. A valid policy was detected for 12,458 websites in August (35.1\%) and for 14,333 websites in October (40.4\%). As we discuss below, the Swiss privacy law revision constitutes a likely driver of the increased policy adoption. All subsequent analyses are conducted on the detected policies within each respective snapshot.

The prevalence of detected policies increases between August and October in all groups (see Figure~\ref{fig:dataset-construction}). However, this increase is strikingly more pronounced in the Swiss-facing groups (\texttt{CH} and \texttt{CH \& EU}) than in the \texttt{EU} group. This suggests differential dynamics around the reform period, indicating that the revised FADP may have driven some Swiss-facing websites without a policy to create one in order to comply with the FADP's disclosure requirements.

We next examine policy length as a coarse proxy for textual changes. Median word counts remain nearly constant in the \texttt{EU} group (+14 words), while policies in \texttt{CH} and \texttt{CH \& EU} increase by approximately a median of 300 words (see Table~\ref{tab:policy-length} in Appendix~\ref{app:policy_length}). The relative increase is marginally larger in the \texttt{CH} group.

\begin{table}[ht]
\centering
\setlength{\tabcolsep}{3pt}
\begin{tabular}{lrrr}
\toprule
 & EU & CH & CH \& EU \\
\midrule
\% GDPR August   & 75.64 & 62.48 & 60.71 \\
\% GDPR October  & 75.09 & 60.77 & 60.75 \\
\midrule
\% FADP August   & 7.59  & 35.83 & 37.10 \\
\% FADP October  & 7.26  & 41.12 & 41.97 \\
\midrule
\textbf{Difference GDPR ($\Delta$ p.p.)} & -0.54 & -1.71 & +0.04 \\
\textbf{Difference FADP ($\Delta$ p.p.)} & -0.33 & +5.29 & +4.88 \\
\bottomrule
\end{tabular}
\caption{Mentions of the GDPR and FADP.}
\label{tab:mentions-gdpr-fadp}
\end{table}

Finally, we analyze explicit references to the GDPR and the FADP in policy texts (Table~\ref{tab:mentions-gdpr-fadp}). To detect such mentions, we compiled multilingual term lists for both laws (e.g., ``DSGVO'' for the GDPR, or ``235.1,'' the FADP’s official compilation number; see Appendix~\ref{app:laws-terms} for the full list) and matched them against each policy. Across all groups, references to the GDPR are substantially more frequent than references to the FADP, including within Swiss-facing websites. Between August and October, mentions of the FADP increase in both Swiss-facing groups, whereas GDPR mentions remain stable.

\subsection{Disclosure Practices} \label{sec:disclosure_practices}
\subsubsection{Snapshot-Level Patterns}

We only examine whether a policy explicitly discloses an item as defined in our codebook. We neither assess full legal validity and enforceability of the policies nor the underlying website processing practices. Our analysis is thus confined to observed disclosures in policy texts. For each policy, we first apply the automated classification procedure described in Section~\ref{sec:method} to determine whether the policy explicitly discloses each obligation defined in our codebook (Section~\ref{sec:codebook}). This produces a binary indicator capturing the presence or absence of the respective disclosure for every obligation and policy snapshot. We then compute the proportion of policies disclosing the respective obligation in August and October, and derive percentage-point differences between the two snapshots.

This yields Table~\ref{tab:disclosures-aug-oct}, which reports disclosure rates in August and October 2023 for each obligation and regulatory exposure group (\texttt{EU}, \texttt{CH}, and \texttt{CH \& EU}), together with percentage-point differences between snapshots. Values are computed separately for each snapshot using all valid policies observed at that time.

\setlength{\tabcolsep}{5pt} 
\begin{table*}[ht]
\centering
\begin{tabular}{lccccccccc}
\toprule
 & \multicolumn{3}{c}{EU} & \multicolumn{3}{c}{CH} & \multicolumn{3}{c}{CH \& EU} \\
\cmidrule(lr){2-4} \cmidrule(lr){5-7} \cmidrule(lr){8-10}
Obligation & Aug & Oct & $\Delta$ p.p.\textsuperscript{1} & Aug & Oct & $\Delta$ p.p.\textsuperscript{1} & Aug & Oct & $\Delta$ p.p.\textsuperscript{1} \\
\midrule
\textit{contr} & 85.8\% & 86.5\% & +0.7 & 73.0\% & 76.9\% & +3.9 & 80.5\% & 83.2\% & +2.7 \\
\textit{purp}  & 99.3\% & 99.3\% & +0.0 & 98.7\% & 98.9\% & +0.2 & 98.4\% & 98.4\% & +0.0 \\
rect  & 87.7\% & 87.5\% & -0.2 & 74.6\% & 77.3\% & +2.7 & 79.9\% & 82.5\% & +2.7 \\
forg  & 89.7\% & 89.3\% & -0.4 & 76.8\% & 79.3\% & +2.5 & 82.5\% & 84.0\% & +1.6 \\
port  & 72.3\% & 71.6\% & -0.6 & 46.4\% & 53.2\% & +6.7 & 54.4\% & 60.0\% & +5.6 \\
comp  & 73.8\% & 73.0\% & -0.8 & 44.0\% & 50.2\% & +6.1 & 52.8\% & 58.0\% & +5.2 \\
\textit{hum}   & 26.6\% & 27.1\% & +0.5 & 16.4\% & 18.7\% & +2.4 & 23.2\% & 24.4\% & +1.2 \\
\midrule
\textbf{Average} & 76.4\% & 76.3\% & -0.1 & 61.4\% & 64.9\% & +3.5 & 67.4\% & 70.1\% & +2.7 \\
\textbf{Total policies} & 2081 & 2176 & -- & 7002 & 8195 & -- & 3375 & 3962 & -- \\
\bottomrule
\end{tabular}
\medskip

\footnotesize{
1. Percentage-point differences ($\Delta$ p.p.) are computed within each group using all valid policies observed in August and October 2023, respectively. \\
The composition of policies differs slightly between snapshots.
}
\vspace{5pt}
\caption{Disclosures by group and obligation (August 2023 versus October 2023).}
\label{tab:disclosures-aug-oct}
\end{table*}

The descriptive evidence reveals marked differences across regulatory exposure groups. The \texttt{EU} group remains largely stable between August and October, with differences for all obligations confined to a narrow range (-0.8 to +0.7 p.p.). In contrast, both \texttt{CH} and \texttt{CH \& EU} groups exhibit consistent positive changes across most obligations. For the \texttt{CH} group, increases range from +2.4 to +6.7 p.p. for all obligations except \texttt{purp}, with the largest gains observed for \texttt{port} and \texttt{comp}. A similar but slightly attenuated pattern is present for \texttt{CH \& EU} websites. Absolute increases tend to be smaller for obligations with higher baseline disclosure levels in August, consistent with ceiling effects in near-saturated categories such as \texttt{purp}.

Baseline disclosure in August is also systematically higher for the \texttt{CH \& EU} group than for \texttt{CH} across nearly all obligations, suggesting prior alignment with GDPR disclosure requirements among jointly exposed websites. By contrast, \texttt{CH}-only websites start from lower baseline levels and exhibit larger absolute increases by October.

To clarify whether these aggregate patterns are driven by particular website subsets, we further disaggregate October disclosure rates by policy update status, website popularity, and policy language. Policies whose text changed between August and October exhibit the highest average disclosure rate (76.0\%), followed by policies that became valid only in October (70.4\%), while unchanged policies disclose the least (64.9\%).\footnote{We provide all details in Table~\ref{tab:oct-policy-status} from Appendix~\ref{app:heterogeneity}.} Website popularity is also associated with higher disclosure completeness: websites appearing in a country-specific Top 5k CrUX bucket average 74.9\%, compared to 67.5\% for less visited websites. Language differences are more limited between German and English policies (68.4\% vs.\ 65.7\%), while Italian policies disclose more (70.5\%) and French policies less (53.3\%); the latter estimates should, however, be interpreted cautiously due to smaller sample sizes. We report the corresponding obligation-level results in Appendix~\ref{app:heterogeneity}.

Overall, the snapshot-level evidence points to disclosure increases concentrated among Swiss-facing websites. Descriptive comparisons alone, however, cannot establish whether these temporal differences reflect statistically meaningful divergence across the three groups.

\subsubsection{Differential Change by Regulatory Exposure}
To formally test whether the descriptive differences documented above reflect systematic divergence across regulatory exposure groups, we conduct an inferential analysis that isolates within-website temporal change from compositional effects. Specifically, we restrict the analysis to the balanced panel of websites with an observed policy in both snapshots. The balanced panel comprises $N = 11{,}800$ websites observed in both August and October 2023, including $2{,}012$ \texttt{EU}, $6{,}682$ \texttt{CH}, and $3{,}106$ \texttt{CH \& EU} websites.\footnote{Group membership is defined according to baseline (August) regulatory exposure.} Restricting the analysis to this panel ensures that estimated changes reflect within-website temporal variation rather than shifts in sample composition.

The central question is whether the change in disclosure rates between August and October is significantly larger for the two groups of Swiss-facing websites compared to the EU quasi-control group. To address this question, we estimate, for each disclosure obligation, a difference-in-differences model on the balanced panel using Ordinary Least Squares (OLS). Given the binary disclosure outcome, this specification operates in probability space, allowing the interaction terms to be interpreted directly as differential percentage-point changes. In contrast, non-linear models for limited dependent variables parameterize interaction effects on the odds scale rather than in levels~\cite{AiNorton2024DiDReview}. We confirm that our substantive conclusions do not depend on the regression model choice by estimating a logistic difference-in-differences model in Appendix~\ref{app:logit_did}.

The dependent variable indicates whether the respective obligation is disclosed in a given policy–snapshot observation. The model includes indicators for time (October vs.\ August), group membership (\texttt{EU} as the reference), and interaction terms between time and group. In addition, we include two control variables to account for potential variation in the baselines: website popularity (CrUX popularity rank buckets) and policy language (German, French, Italian, or English). These controls account for systematic differences in disclosure practices associated with market reach and linguistic context rather than the Swiss privacy law revision. Standard errors are clustered at the website level to account for repeated observations of the same policy across snapshots.

In this linear specification, the interaction coefficients can be interpreted directly as differential percentage-point changes in disclosure rates relative to the EU baseline. As in any difference-in-differences design, identification relies on the assumption that, absent the Swiss regulatory revision, temporal changes in disclosure would not have differed systematically between Swiss-facing and EU websites. While the two-snapshot structure does not allow direct testing of parallel pre-trends, the empirical stability observed in the \texttt{EU} group between August and October provides a plausible baseline against which differential changes can be evaluated. Moreover, over short time intervals, policies can reasonably be expected to remain stable in the absence of a relevant event, such as a court case or extensive news coverage.

Because we estimate separate models for multiple disclosure obligations, conducting parallel hypothesis tests increases the risk of false positives. We, therefore, control the expected false discovery rate using the Benjamini--Hochberg~\cite{Benjamini1995FDR} procedure at $\alpha = 0.05$, applied separately to the two families of interaction terms (\texttt{CH} vs.\ \texttt{EU} and \texttt{CH \& EU} vs.\ \texttt{EU}).

Table~\ref{tab:did-results} reports the estimated percentage-point changes within each group alongside the difference-in-differences interaction coefficients relative to the EU baseline. The \texttt{EU} group exhibits minimal changes across obligations, supporting the assumption that policies remain largely stable over short time horizons in the absence of a regulatory change. By contrast, the interaction terms for the \texttt{CH} and \texttt{CH \& EU} groups are positive and statistically significant for all obligations except \texttt{purp}, indicating that Swiss-facing websites experience significantly larger increases over time than the EU quasi-control group. The strongest differential effects are observed for \texttt{port} and \texttt{comp}, which also display the largest absolute increases in percentage-point terms (see Table~\ref{tab:disclosures-aug-oct}).

\begin{table*}[ht]
\centering
\setlength{\tabcolsep}{4pt}
\begin{tabular}{lccccc}
\toprule
 & \multicolumn{3}{c}{Temporal Change ($\Delta$ p.p.)} 
 & \multicolumn{2}{c}{Differential Effect (vs.\ EU)} \\
\cmidrule(lr){2-4} \cmidrule(lr){5-6}
Obligation 
 & EU 
 & CH 
 & CH \& EU 
 & CH: DiD Coef.\ (95\% CI) 
 & CH \& EU: DiD Coef.\ (95\% CI) \\
\midrule
\textit{contr} & +0.35 & +2.38 & +1.96 & 2.03$^{*}$ (1.53--2.53) & 1.62$^{*}$ (0.98--2.26) \\
\textit{purp}  & +0.10 & -0.03 & -0.13 & -0.13 (-0.31--0.06)      & -0.23 (-0.46--0.01) \\
rect  & +0.20 & +2.05 & +2.38 & 1.85$^{*}$ (1.34--2.36) & 2.18$^{*}$ (1.46--2.91) \\
forg  & +0.15 & +2.17 & +1.71 & 2.02$^{*}$ (1.51--2.53) & 1.56$^{*}$ (0.85--2.27) \\
port  & +0.10 & +4.89 & +4.93 & 4.79$^{*}$ (4.15--5.44) & 4.83$^{*}$ (3.88--5.77) \\
comp  & +0.00 & +4.15 & +4.15 & 4.15$^{*}$ (3.53--4.76) & 4.15$^{*}$ (3.25--5.05) \\
\textit{hum}   & +0.05 & +1.69 & +1.61 & 1.64$^{*}$ (0.99--2.29) & 1.56$^{*}$ (0.68--2.44) \\
\bottomrule
\end{tabular}

\medskip
\footnotesize{
Percentage-point changes ($\Delta$ p.p.) are computed on the balanced panel of websites observed in both snapshots ($N = 11{,}800$; $23{,}600$ policy--snapshot observations). 

$^{*}$ indicates statistical significance after Benjamini--Hochberg~\cite{Benjamini1995FDR} correction at $\alpha = 0.05$, applied separately to the CH vs.\ EU and CH \& EU vs.\ EU interaction effects.
}
\caption{Differential change in disclosure rates between August and October 2023.}
\label{tab:did-results}
\end{table*}

We further examine whether these differential changes are concentrated among particular website subsets by estimating interaction models with an aggregate disclosure index (Appendix~\ref{app:heterogeneity}, Table~\ref{tab:heterogeneity-interactions}). The results indicate that website popularity primarily explains disclosure levels: Top~5k websites disclose more in October, but the differential August--October increase is not confined to them. Language mediates the response more substantially. The increase is strongest among German-language policies, which dominate the Swiss-facing sample, while estimates for policies in other languages are smaller and should thus be interpreted with caution. Finally, a descriptive decomposition shows that within-website disclosure changes are concentrated among policies whose text changed between snapshots, consistent with active policy revision during the observation window.

Taken together, the inferential analysis shows that the temporal increases are concentrated among websites directly exposed to the Swiss privacy law revision and significantly exceed the minor changes observed in the EU control group. While alternative explanations, such as other concurrent jurisdiction-specific developments, cannot be entirely excluded, the observed pattern is consistent with a significant increase in both mandatory (according to the revised Swiss privacy law) and voluntary transparency disclosures in the policies of Swiss-facing websites.

\section{Privacy Policy Generators}\label{sec:generators}

As website operators pondered how to best update their policies to become compliant with Swiss and potentially also EU privacy law, our data offers preliminary evidence that some turned to automated policy generators (\textit{generators}). In the following sections, we describe the key characteristics of generators and the method we used to detect their use to draft policies. We find that generators are not only explicitly used by many Swiss-facing websites (18.2\% for October policies in the \texttt{CH} group), but that their use (1) slightly increases over time and (2) is correlated with increases in disclosure rates by up to 15 p.p. for some transparency obligations. We also demonstrate that generators tend to output similar policies with identical clauses, which likely explains the consistently high disclosure rates exhibited by some generators.

\subsection{Characteristics}\label{sec:generators-characteristics}

While the various business models and functionalities of generators complicate any simple definition, the core principle of a generator is to \textit{guide users through a questionnaire or similar process that ultimately results in a tailored and (almost) ready-made policy} (see generally~\cite{betts-dawn-2017, contreras-solving-2025, foster_when_2018}). We display part of the translated user interface for the most widely used (yet no longer operational) generator in our dataset called ``SwissAnwalt'' in Appendix~\ref{app:swissanwalt}. Importantly, users had to agree to cite SwissAnwalt as the source of their policy under threat of litigation before their tailored policy text was generated. We are unclear whether SwissAnwalt actually followed through on litigation threats, but evidence exists that a German law firm offering a generator sends warning letters when its generator is not referenced in the policy~\cite{steiger-deutscher-2022}.

Other generators achieve an even higher level of automation, essentially skipping the error-prone, manual entry of information by the user. For example, PrivacyBee~\cite{privacybee-datenschutz-2025} advertises a policy generator on ``autopilot,'' claiming to automatically recognize all privacy-relevant services used on a given website and to generate as well as regularly update the website's policy. For this service, PrivacyBee charges a yearly fee of around \$70 per domain. 

None of the generators we identify in our study (see Table~\ref{tab:generator-prevalence} below) publicly mention the use of generative AI to draft policies. While this may change in the future and our definition does not preclude it, we assume that policy texts are standardized enough for generators to rely entirely on pre-defined clauses that are combined together depending on the user input. Note that we exclude mere templates that users must manually complete or modify from our definition of generators, as generators are based on the notion that the \textit{user is not intended to directly interact with the actual text of the policy}. While we carefully and manually ensure that all generators identified in this study fulfill this definition (see Section~\ref{sec:generators_identification}), we acknowledge that it is not always possible to perfectly and objectively distinguish a generator from a template.

\subsection{Generator Identification Method} \label{sec:generators_identification}

To detect the prevalence of generators, we first sought to identify all major providers. For this purpose, we introduced the exploratory question \texttt{org} in our LLM-based analysis: ``Does the policy mention that it has been created with the help of a tool (e.g., policy generator) or template?'' This very broad definition was used to collect potential generators inclusively, with the aim of discarding false positives in a subsequent manual step. Unlike the nine disclosure dimensions defined in Section~\ref{sec:method}, the response to \texttt{org} is a string value representing the name of the stated candidate generator, and we did not validate the model's output due to its exploratory nature.

Next, we manually inspected the model output to remove false positives. Using the identified generators and associated text in policies, we then created a reliable, conservative dictionary of RegEx patterns that we used to detect acknowledgment of each generator in all policies (see artifact repository). We refined the RegEx patterns until we achieved perfect accuracy on random samples of ten policies per generator. We manually verified that the resulting 140 policies all actually disclosed the relevant generator. As a result, precision is high, and our estimates should be interpreted as a conservative lower bound of generator use. This procedure allowed us to overcome the model's challenge of recognizing \textit{ex ante} whether the stated source of a policy is a generator or a mere template.

\subsection{Generator Use and Disclosures} \label{sec:generators_use}

Table \ref{tab:generator-prevalence} presents generator use statistics. Generator use is pervasive, with over 2,000 policies explicitly produced by one in the October dataset. Their use accounts for 18.2\% of all policies in the October \texttt{CH} group. We observe disproportionately higher use of generators for Swiss-facing websites. We also found use of EU generators for Swiss-facing websites, but \textit{not} the opposite.

\begin{table}[ht]
\small
\centering
\begin{threeparttable}
\begin{tabular}{lcccccc}
\toprule
 & \multicolumn{2}{c}{EU} & \multicolumn{2}{c}{CH} & \multicolumn{2}{c}{CH \& EU} \\
\cmidrule(lr){2-3} \cmidrule(lr){4-5} \cmidrule(lr){6-7}
Generator\tnote{1} & Aug & Oct & Aug & Oct & Aug & Oct \\
\midrule
SwissAnwalt (CH) & 0 & 0 & 549 & 539 & 137 & 123 \\
Datenschutzpartner (CH) & 0 & 0 & 104 & 239 & 46 & 98 \\
eRecht24 (DE) & 9 & 9 & 146 & 136 & 11 & 11 \\
PrivacyBee (CH) & 0 & 0 & 55 & 177 & 20 & 45 \\
DGD (DE) & 28 & 29 & 64 & 61 & 40 & 34 \\
activeMind (DE) & 21 & 21 & 63 & 62 & 16 & 15 \\
BrainBox (CH) & 0 & 0 & 49 & 96 & 16 & 29 \\
Schwenke (DE) & 17 & 17 & 43 & 39 & 24 & 23 \\
AdSimple (DE) & 9 & 9 & 41 & 43 & 5 & 8 \\
WeissPartner (DE) & 5 & 5 & 32 & 30 & 12 & 12 \\
Iubenda (IT) & 5 & 5 & 20 & 23 & 11 & 15 \\
MeinDatenschutz (DE) & 9 & 10 & 16 & 16 & 5 & 5 \\
LegallyOK (CH) & 0 & 0 & 6 & 32 & 1 & 7 \\
TrustedShops (DE) & 0 & 0 & 14 & 3 & 5 & 5 \\
\midrule
\textbf{With generator}\tnote{2} & 102 & 104 & 1200 & 1492 & 347 & 424 \\
\textbf{\% of all policies} & 4.9 & 4.8 & 17.1 & 18.2 & 10.3 & 10.7 \\
\bottomrule
\end{tabular}
\begin{tablenotes}
\item[1] Name with country of origin in parenthesis.
\item[2] Policies stating having used \textit{at least one} generator. Only 13 policies across all groups state having used two or more generators.
\end{tablenotes}
\caption{Prevalence of generators (sorted by total use).}
\label{tab:generator-prevalence}
\end{threeparttable}
\end{table}

In Table \ref{tab:generator-by-language-and-rank}, we show generator use by policy language and top website rank according to CrUX, i.e., the highest popularity bucket across all countries (see Section~\ref{sec:dataset}), for all October policies. In this table and in all the following ones, we analyze the policies of all three groups together. Generators have predominantly been used for German policies (16.4\%), with all other languages averaging less than 4\%. This is likely due to many generators only being available in German, which may explain the discrepancy in generator use between Swiss- and EU-facing websites, as the latter group contains a smaller relative proportion of German-language policies (see Appendix~\ref{app:languages}, Table~\ref{tab:languages}).

\setlength{\tabcolsep}{3pt}
\begin{table}[ht]
\centering
\begin{tabular}{lrrrrrrr}
\toprule
 & \multicolumn{4}{c}{Language} & \multicolumn{3}{c}{Top Website Rank} \\
\cmidrule(lr){2-5} \cmidrule(lr){6-8}
 & DE & EN & IT & FR & 5k & 10--50k & 100k+ \\
\midrule
Total policies & 11855 & 1600 & 725 & 153 & 1108 & 7224 & 6001 \\
With generator & 1945 & 62 & 13 & 0 & 47 & 1000 & 973 \\
\midrule
\textbf{Percentage} & 16.4\% & 3.9\% & 1.8\% & 0.0\% & 4.2\% & 13.8\% & 16.2\% \\
\bottomrule
\end{tabular}
\caption{Prevalence of generators by policy language and website popularity rank (all groups, October 2023).}
\label{tab:generator-by-language-and-rank}
\end{table}

Moreover, generators are mostly used by less popular websites, with less than 5\% of the policies from Top 5k websites indicating use. This intuitively makes sense: larger websites are more exposed to legal and reputational risks, and they presumably have more resources to hire internal or external legal advisors.

We also inspected how generator use affects the disclosures of policies. Table~\ref{tab:generator-disclosures} illustrates that policies created with a generator are associated with increased disclosure rates of up to 15 p.p. (right to lodge a complaint). Disclosure of the controller's identity---required by both Swiss and EU privacy law---is also noticeably higher (+9.5 p.p.). This pattern holds even when disaggregating generator use by website popularity rank (see Appendix~\ref{app:generator-details}).

The only exception is automated decisions, which are mentioned by 7.5 p.p. fewer policies with a generator. This could be due to the fact that generators are mostly used by smaller websites (see Table~\ref{tab:generator-by-language-and-rank}), which perform automated decision-making less frequently than larger websites (Top 5k websites refer to automated decisions 13.2 p.p. more than non-Top 5k websites, see Appendix~\ref{app:heterogeneity}, Table~\ref{tab:disclosure-details}). Without knowing which websites actually rely on automated decisions, we lack a conclusive explanation.

\setlength{\tabcolsep}{5pt}
\begin{table}[ht]
\centering
\begin{tabular}{lccc}
\toprule
 & \multicolumn{3}{c}{Generator Used} \\
\cmidrule(lr){2-4}
Obligation & No & Yes & $\Delta$ p.p.\textsuperscript{1} \\
\midrule
\textit{contr} &    78.8\% &   88.3\% & + 9.5 \\
\textit{purp} &    98.6\% &  100.0\% & + 1.3 \\
rect &    79.7\% &   83.8\% & + 4.0 \\
forg &    81.8\% &   84.0\% & + 2.1 \\
port &    56.0\% &   69.0\% & +13.0 \\
comp &    53.7\% &   68.7\% & +15.0 \\
\textit{hum} &    22.6\% &   15.1\% & -7.5 \\
\midrule
\textbf{Average} &    67.3\% &   72.7\% & + 5.3 \\
\textbf{Total policies} & 12313 & 2020\textsuperscript{2} & -- \\
\bottomrule
\end{tabular}
\medskip

\footnotesize{1. The $\Delta$ column reports percentage-point differences for generator use. \\
2. October policies stating having used \textit{at least one} generator.}
\caption{Disclosures by generator use (October 2023).}
\label{tab:generator-disclosures}
\end{table}

Finally, we inspected the heterogeneous disclosure across all individual generators in our sample and present our findings for the Top 5 generators in Table~\ref{tab:top5-generator-obligation-single} (see Appendix~\ref{app:generator-details} for the full list). While most generators achieve higher average disclosure rates than the 67.3\% observed for policies without any generator use (Table~ \ref{tab:generator-disclosures}), there are two prominent exceptions: SwissAnwalt and eRecht24.

\begin{table*}[ht]
\centering
\begin{tabular}{l|rrrrrrr|rrr}
\toprule
Generator & \textit{contr} & \textit{purp} & rect & forg & port & comp & \textit{hum} & \textbf{Average} & \textbf{Policies} & \textbf{Market Share} \\
\midrule
SwissAnwalt (CH) & 92.5\% & 100.0\% & 57.8\% & 58.3\% & 45.5\% & 46.1\% & 1.4\% & 57.4\% &   657 & 32.7\% \\
Datenschutzpartner (CH) & 99.1\% & 99.7\% & 99.7\% & 99.7\% & 89.3\% & 99.7\% & 0.6\% & 84.0\% &   337 & 16.7\% \\
PrivacyBee (CH) & 99.5\% & 100.0\% & 99.5\% & 99.5\% & 98.6\% & 99.5\% & 47.7\% & 92.1\% &   222 & 11.0\% \\
eRecht24 (DE) & 5.8\% & 100.0\% & 98.1\% & 98.1\% & 5.8\% & 5.8\% & 1.3\% & 45.0\% &   156 & 7.8\% \\
DGD (DE) & 98.3\% & 100.0\% & 98.3\% & 98.3\% & 97.5\% & 95.0\% & 99.2\% & 98.1\% &   121 & 6.0\% \\
\bottomrule
\end{tabular}
\medskip

\caption{Disclosure rates by Top 5 generator and obligation for policies using a single generator (October 2023).}
\label{tab:top5-generator-obligation-single}
\end{table*}

SwissAnwalt illustrates the impact of choice architecture and default configurations used by generators. As shown in Figure~\ref{fig:swissanwalt}, the generator provided an option to include data subject rights in the resulting policy. Since Swiss data protection law does not mandate the disclosure of these rights, the corresponding selection box was \textit{not} pre‑checked. Consequently, data subject rights appeared in the generated output only when users explicitly opted in. Although a substantial share of SwissAnwalt users nonetheless chose to disclose these rights, many did not. Given the high disclosure rates of the other two major Swiss generators, Datenschutzpartner and PrivacyBee, we suspect they mentioned these rights by default.

The story with eRecht24 is different: policies generated with the tool simultaneously exhibit very high values (like the right of access and rectification) and very low values (like the right to data portability). Having inspected all October policies referencing eRecht24, we found the following exact boilerplate text in 92.3\% of them (translated from German):

\begin{quote}

    \textit{You have the right to receive information for free at any time about your stored personal data, its origin and recipients, and the purpose of data processing, as well as the right to rectify, block, or delete this data.}
    
\end{quote}

The text does not mention the rights to data portability and to lodge a complaint. Even though eRecht24 has been providing a checklist that states that these two rights must be mentioned in the policy since at least 2022~\cite{siebert-dsgvo-konforme-2025}, most websites have apparently not updated their policies accordingly. This is particularly salient given that eRecht24 is based in Germany, where the GDPR requires websites to disclose all data subject rights.

\subsection{Standardization and Clusters} \label{sec:generators_similarity}

The use of standard clauses for disclosures is not limited to eRecht24: the vast majority (88.3\%) of policies generated with PrivacyBee---which exhibit very high transparency disclosures---contain the exact same paragraph disclosing user rights. This shows how textual standardization can be either beneficial or detrimental depending on the quality of the standardized clause. It also demonstrates the path dependencies of generated policies if their pre-defined clauses are not updated over time. For the generator Datenschutzpartner, by contrast, we found evidence for several different versions of the transparency disclosures.

These observations motivated us to systematically analyze the extent to which generators output standardized policies. This matters both in terms of disclosures mandated by Swiss privacy law as well as the diffusion of the GDPR on a voluntary basis, as generators with standardized clauses exhibiting a high market share can significantly affect both dimensions. In short, we find evidence consistent with a relatively high degree of standardization.

We assess standardization in generator-produced policies by embedding all German-language policies from the October snapshot and projecting them to two dimensions using t-SNE~\cite{maaten-visualizing-2008} based on cosine similarity. Similar visualization approaches have been used to analyze standardization in legal documents~\cite{bartlett-standardization-2026,badawi-value-2023}. These traditional approaches first used term-document matrices like TF-IDF to vectorize the text data. By contrast, we used the text embedding model \texttt{text-embedding-3-large} from OpenAI~\cite{openai-new-2024} to preserve the semantic structure of the policies. The embeddings were based on each policy's first 8,192 tokens, due to the model's context window limitation. While some policies exceed this window, Badawi et al.~\cite[p. 16]{badawi-value-2023} demonstrated strong performance clustering legal documents with only the first 1,000 characters. We colored the policies of the ten most widely used generators (see Table~\ref{tab:generator-prevalence}).

\begin{figure*}[ht]
    \centering
    \includegraphics[width=0.75\linewidth]{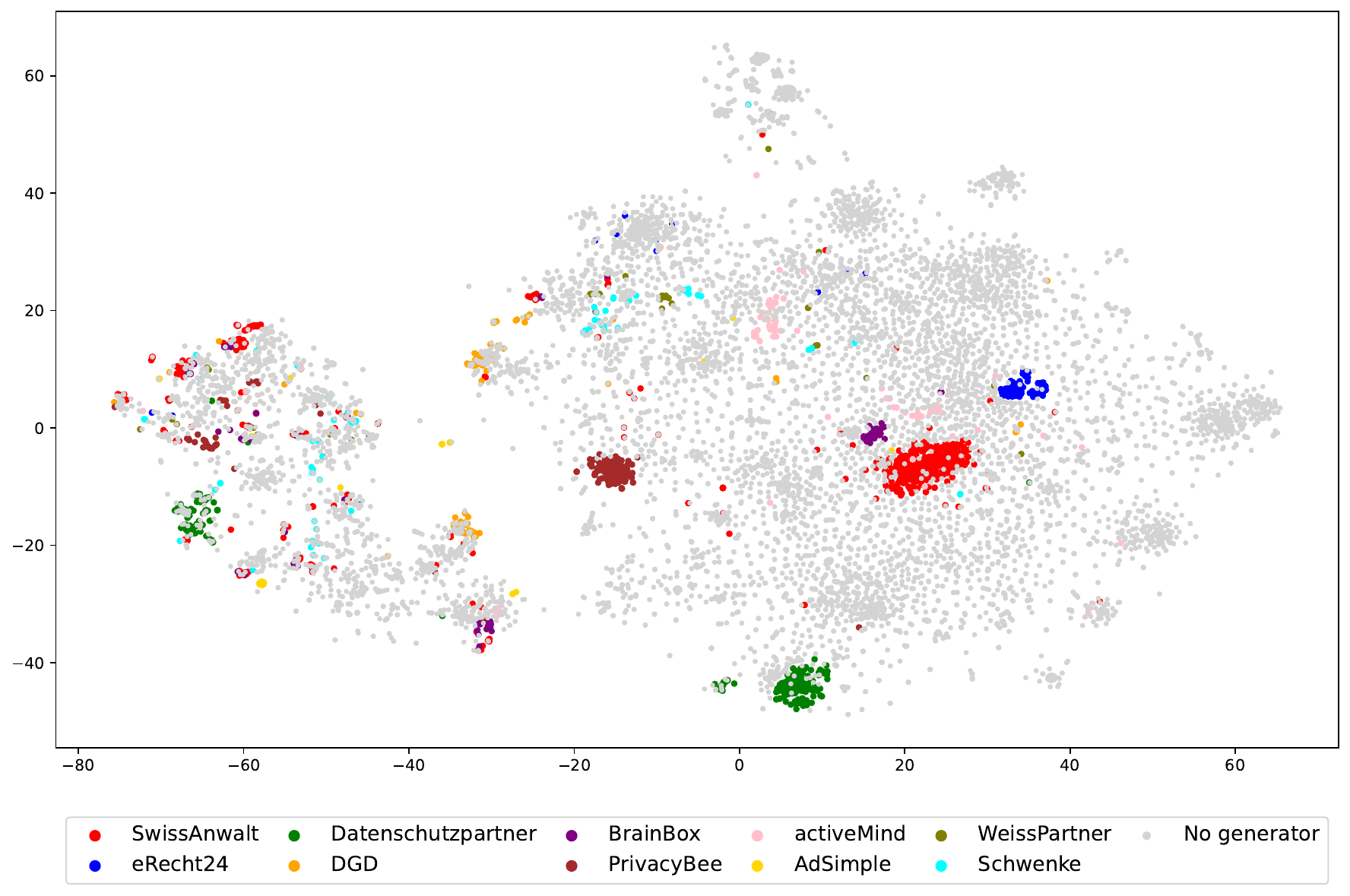}
    \caption{Cluster analysis of policies by generator (Top 10) for October policies in German ($N = 11,855$).}
    \Description{t-SNE representation of all October policies on a two-dimensional space. The visualization illustrates different colors for different generators. The generators form clear, albeit not fully distinct clusters.}
    \label{fig:tsne-generators}
\end{figure*}

As shown in Figure \ref{fig:tsne-generators}, generators form distinct, albeit imperfect, clusters. One possible reason for these imperfect clusters would be that websites modify the generator's output. For instance, one website stated (translated from German):

\textit{\begin{quote}
    This privacy policy was created using the privacy policy generator from DGD [...] with a lawyer specializing in data protection law, and the model privacy policy provided by the law firm Weiß \& Partner~\cite{schneemenschen-gmbh-datenschutzerklarung-2018}.
\end{quote}} 

The provision of different outputs for different data processing practices is, moreover, an inherent feature of generators. Generated policies that fall outside corresponding clusters may simply originate from websites with unusual data practices (e.g., the collection of atypical data). Furthermore, generators evolve over time: different versions of the same generator (e.g., before and after the GDPR) likely coexist in the October corpus of policies.

\section{Discussion} \label{sec:discussion}

\subsection{Assessment and Policy Implications}

Our analysis offers evidence that the revised Swiss privacy law significantly affected Swiss-facing websites' policies. Moreover, we find that the revision led to an increase of voluntary disclosures of data subject rights, which would be consistent with---but not direct evidence of---a delayed Brussels Effect of the GDPR in Switzerland. Taken together, these findings highlight that, by revising their privacy laws, smaller jurisdictions can meaningfully increase both mandatory and voluntary privacy disclosures.

Despite the revised Swiss privacy law having entered into force back in 2023, it remains relevant to this very day. The Swiss data protection authority~\cite{federal_data_protection_and_information_commissioner_data_2025} announced having received 1,406 reports of violations of the Swiss privacy law from its entry into force through January 2025. With Switzerland contemplating whether and how to regulate artificial intelligence, both the Swiss Federal Office of Justice~\cite{federal_office_of_justice_rechtliche_2024} and a prominent legal scholar~\cite{thouvenin_rechtsrahmen_2026} have also proposed to extend the current scope of the right to be informed about automated decision-making (\texttt{hum}).

The generally strong performance of GPT-5 in assessing policy disclosures in our multilingual pipeline raises two noteworthy discussion points. First, our dataset demonstrates how transparency disclosures can be reliably identified in different languages with a single English prompt. This implies that regulators in multilingual jurisdictions can efficiently evaluate policy disclosures in German, French, Italian, and English using our pipeline. The two caveats are (1) that there must be an underlying codebook and annotated dataset, which we release with our study, and (2) that the method does not guarantee perfect performance, particularly beyond the reasonably objective attributes we considered. Nonetheless, we argue that having our method assist regulators in identifying potential non-compliant policies could provide significant benefits. Second, as our performance for the automated decision-making metric \texttt{hum} indicates (see Figure~\ref{fig:f1-gpt5}), LLMs may still struggle to assess specific dimensions of disclosures, which underlines the importance of both carefully drafted codebooks and cautious AI use.

Finally, we find that generators are not only particularly prevalent for smaller German-language websites, but that their use is connected to higher rates of transparency disclosures in the context of this study. Policymakers should take note, as generators need not be for-profit: the UK's Information Commissioner’s Office provides a free policy generator~\cite{information-commissioners-office-ico-create-2024}. We welcome this effort, as our findings suggest that generators can support small business compliance. This is not to say that generators are perfect: Pan et al.~\cite{pan_is_2024} document flaws in the synthetically generated policies they inspect, and we fear that issues in generated policies may not be subsequently rectified (see Section~\ref{sec:generators_use}). Policymakers might wish to consider generators as an intervention point, encouraging routine auditing as well as timely updates of both the system itself and outdated generated policies.

\subsection{Limitations} \label{sec:limitations}

Our study has limitations typical of empirical work in this area. The short temporal time window was deliberately chosen to avoid weakening our causal identification method by measuring unrelated subsequent events. We concede that the reported changes likely constitute lower bounds of the revision's true effect, as some organizations may have opted to comply at a later time. Our grouping procedure, moreover, inevitably oversimplifies the territorial scope of Swiss and European privacy law, but we believe it approximates the websites subject to each privacy law regime reasonably well. Our dataset also focuses on a restricted set of variables. Additional dimensions such as website categories could be used as controls in the regression analysis. 

While manual annotations inevitably involve a degree of subjectivity, we alleviated this issue through the iterative refinement of our codebook until annotators reached at least acceptable agreement. Furthermore, the LLM-based policy analysis cannot guarantee perfect reproducibility, since model outputs may vary slightly across runs. Finally, our conservative generator identification method minimizes the risk of false positives, potentially at the cost of underestimating the true use of generators for drafting policies.

Despite these limitations, our paper sheds light on (1) the real-world effects of a privacy law revision beyond the widely studied European and American privacy laws, (2) the potential of LLMs at large-scale multilingual disclosure assessments, and (3)~the important role that generators play in website policies.

\section{Conclusion and Future Work} \label{sec:conclusion}

This paper extends our understanding of the real-world impact of privacy law changes by investigating the effects of a Swiss privacy law revision for 35,000 websites. Developing and validating a novel multilingual disclosure assessment pipeline for policies, we demonstrate that the revision significantly increased privacy disclosures required by Swiss or European law. We also find that generators can play an important role in helping smaller websites comply with privacy legislation. This work paves the way for future research along the following dimensions and others:

\begin{enumerate}
    \item How do privacy law revisions in different jurisdictions affect transparency disclosures in policies?
    \item Do larger organizations with multilingual or multijurisdictional policies voluntarily implement privacy law revisions across all their policy versions?
    \item To what extent can our multilingual disclosure assessment pipeline be used for different types of legal documents, such as terms of service?
    \item Is generator use equally widespread in other countries and for different legal documents?
    \item What can policymakers draw from these findings to improve the details of privacy laws, monitoring and enforcement, and quality and ease of transparency disclosures?
\end{enumerate}

Pursuing this line of research could yield three key benefits for enhancing privacy: (1) assessing the efficacy of privacy laws from different jurisdictions can help us understand which approaches work and which do not; (2) LLMs may support regulators in monitoring specific aspects of compliance at scale; and (3) generators may assist with privacy-related disclosures, particularly for resource-constrained smaller businesses. We hope our work will spark future research in these directions.


\begin{acks}
The authors used generative AI-based tools to revise the text, improve flow, and correct any typos, grammatical errors, and awkward phrasing. The authors also acknowledge having used AI-based tools for assistance in analyzing the data and editing LaTeX.

The authors would like to thank Elias Landes for his excellent research assistance.

Luka Nenadic gratefully acknowledges the support of the Swiss National Science Foundation (project 10002634).

The work of David Rodriguez was partially supported by the Re-InitS project, funded by the Plan Estatal de Investigación Científica y Técnica y de Innovación 2024–2027 (Ministerio de Ciencia e Investigación (Spain) - MCIN/AEI/10.13039/501100011033) under grant agreement PID2024-155230OB-C43 and by ``DINA-iOS: Desarrollo de una infraestructura experimental para la auditoría de aplicaciones iOS" project funded by ETSI Telecomunicación under ``Ayudas Primeros Proyectos" grant.
\end{acks}

\bibliographystyle{ACM-Reference-Format}
\bibliography{main}

@String{Computing = "Computing" }

@String{Computer = "{IEEE} Computer" }

@String{Springer = "Springer-Verlag" }

@inproceedings{harkous-polisis-2018,
	location = {Baltimore, MD, USA},
	title = {Polisis: {Automated} analysis and presentation of privacy policies using deep learning},
    booktitle = {Proceedings of the 27th USENIX Security Symposium},
	shorttitle = {Polisis},
	url = {https://www.usenix.org/conference/usenixsecurity18/presentation/harkous},
	lastaccessed = {2024-05-24},
	publisher = {USENIX Association},
    address   = {Berkeley, CA, USA},
	author = {Harkous, Hamza and Fawaz, Kassem and Lebret, Rémi and Schaub, Florian and Shin, Kang G. and Aberer, Karl},
	year = {2018},
	pages = {531--548},
    series = {USENIX Security '18}
}

@book{grimmer-text-2022,
	address = {Princeton, NJ, USA},
	title = {Text as {Data}: {A} {New} {Framework} for {Machine} {Learning} and the {Social} {Sciences}},
	shorttitle = {Text as {Data}},
	publisher = {Princeton University Press},
	author = {Grimmer, Justin and Roberts, Margaret E. and Stewart, Brandon M.},
	year = {2022},
}

@inproceedings{wilson-creation-2016,
	location = {Berlin, Germany},
	title = {The creation and analysis of a website privacy policy corpus},
	doi = {10.18653/v1/P16-1126},
	booktitle = {Proceedings of the 54th {Annual} {Meeting} of the {Association} for {Computational} {Linguistics}},
	publisher = {Association for Computational Linguistics},
    address = {Stroudsburg, PA, USA},
	author = {Wilson, Shomir and Schaub, Florian and Dara, Aswarth Abhilash and Liu, Frederick and Cherivirala, Sushain and Giovanni Leon, Pedro and Schaarup Andersen, Mads and Zimmeck, Sebastian and Sathyendra, Kanthashree Mysore and Russell, N. Cameron and B. Norton, Thomas and Hovy, Eduard and Reidenberg, Joel and Sadeh, Norman},
	year = {2016},
	series = {ACL '16},
	pages = {1330--1340},
}

@inproceedings{amos-privacy-2021,
	location = {Ljubljana, Slovenia},
	title = {Privacy policies over time: {Curation} and analysis of a million-document dataset},
	shorttitle = {Privacy {Policies} over {Time}},
	doi = {10.1145/3442381.3450048},
	booktitle = {Proceedings of the {Web} {Conference} 2021},
	publisher = {Association for Computing Machinery},
    address = {Stroudsburg, PA, USA},
	author = {Amos, Ryan and Acar, Gunes and Lucherini, Eli and Kshirsagar, Mihir and Narayanan, Arvind and Mayer, Jonathan},
	month = apr,
	year = {2021},
	pages = {2165--2176},
    series = {WWW '21}
}

@inproceedings{arora-tale-2022,
	location = {Marseille, France},
	title = {A tale of two regulatory regimes: {Creation} and analysis of a bilingual privacy policy corpus},
	shorttitle = {A {Tale} of {Two} {Regulatory} {Regimes}},
	url = {https://aclanthology.org/2022.lrec-1.585},
	lastaccessed = {2024-05-10},
	booktitle = {Proceedings of the {Thirteenth} {Language} {Resources} and {Evaluation} {Conference}},
	author = {Arora, Siddhant and Hosseini, Henry and Utz, Christine and Bannihatti Kumar, Vinayshekhar and Dhellemmes, Tristan and Ravichander, Abhilasha and Story, Peter and Mangat, Jasmine and Chen, Rex and Degeling, Martin and Norton, Thomas and Hupperich, Thomas and Wilson, Shomir and Sadeh, Norman},
	editor = {Calzolari, Nicoletta and Béchet, Frédéric and Blache, Philippe and Choukri, Khalid and Cieri, Christopher and Declerck, Thierry and Goggi, Sara and Isahara, Hitoshi and Maegaard, Bente and Mariani, Joseph and Mazo, Hélène and Odijk, Jan and Piperidis, Stelios},
    publisher = {Association for Computational Linguistics},
    address = {Stroudsburg, PA, USA},
	year = {2022},
	series = {LREC '22},
	pages = {5460--5472},
}

@article{peukert-regulatory-2022,
	title = {Regulatory spillovers and data governance: {Evidence} from the {GDPR}},
	volume = {41},
	doi = {10.1287/mksc.2021.1339},
	number = {4},
	journal = {Marketing Science},
	author = {Peukert, Christian and Bechtold, Stefan and Batikas, Michail and Kretschmer, Tobias},
	month = feb,
	year = {2022},
	pages = {746--768},
}

@article{linden-privacy-2020,
	title = {The privacy policy landscape after the {GDPR}},
	volume = {2020},
	doi = {https://doi.org/10.2478/popets-2020-0004},
	number = {1},
	journal = {Proceedings on Privacy Enhancing Technologies},
	author = {Linden, Thomas and Khandelwal, Rishabh and Harkous, Hamza and Fawaz, Kassem},
	year = {2020},
	pages = {47--64},
}

@inproceedings{dabrowski-measuring-2019,
	address = {Cham},
	title = {Measuring cookies and web privacy in a post-{GDPR} world},
	doi = {https://doi.org/10.1007/978-3-030-15986-3-17},
	booktitle = {Passive and {Active} {Measurement}},
	publisher = {Springer},
	author = {Dabrowski, Adrian and Merzdovnik, Georg and Ullrich, Johanna and Sendera, Gerald and Weippl, Edgar},
	editor = {Choffnes, David and Barcellos, Marinho},
	year = {2019},
	series = {PAM '19},
	pages = {258--270},
}

@article{kahan-standardization-1997,
	title = {Standardization and innovation in corporate contracting (or "the economics of boilerplate")},
	volume = {83},
	doi = {10.2307/1073747},
	number = {4},
	journal = {Virginia Law Review},
	author = {Kahan, Marcel and Klausner, Michael},
	month = may,
	year = {1997},
	pages = {713--770},
}

@article{frankenreiter-cost-based-2022,
	title = {Cost-based {California} {Effects}},
	volume = {39},
	url = {https://www.yalejreg.com/print/cost-based-california-effects/},
    lastaccessed = {2025-05-25},
	journal = {Yale Journal on Regulation},
	author = {Frankenreiter, Jens},
	year = {2022},
	pages = {1155--1217},
}

@article{davis-filling-2024,
	title = {Filling the void: {How} {E}.{U}. privacy law spills over to the {U}.{S}.},
	volume = {1},
	doi = {10.1177/2755323X241237619},
	journal = {Journal of Law and Empirical Analysis},
	author = {Davis, Kevin E. and Marotta‐Wurgler, Florencia},
	month = jun,
	year = {2024},
	pages = {1--21},
}

@book{bradford-brussels-2020,
	address = {Oxford, United Kingdom},
	title = {The {Brussels} {Effect}},
	publisher = {Oxford University Press},
	author = {Bradford, Anu},
	year = {2020},
}

@article{bartlett-standardization-2026,
author = {Bartlett, Robert},
title = {Standardization and innovation in venture capital contracting: {Evidence} from startup company charters},
journal = {The Journal of Legal Studies},
volume = {55},
number = {1},
pages = {83-128},
year = {2026},
doi = {10.1086/736135},
}

@misc{badawi-value-2023,
	title = {The value of {M}\&{A} drafting},
	url = {https://ssrn.com/abstract=4337075},
	language = {en},
	author = {Badawi, Adam B. and de Fontenay, Elisabeth and Nyarko, Julian},
	year = {2023},
    archivePrefix = {{SSRN} {Working} {Paper}},
    eprint        = {4337075},
}

@inproceedings{rodriguez-data-2024,
	title = {Data retention disclosures in the {Google} {Play} {Store}: {Opacity} remains the norm},
    booktitle = {2024 IEEE European Symposium on Security and Privacy Workshops},
	shorttitle = {Data {Retention} {Disclosures} in the {Google} {Play} {Store}},
	doi = {10.1109/EuroSPW61312.2024.00009},
	author = {Rodriguez, David and Fernández-Aller, Celia and Del Alamo, Jose M. and Sadeh, Norman},
	month = jul,
	year = {2024},
    address = {Vienna, Austria},
    publisher = {IEEE},
	pages = {19--23},
    series = {EuroS\&PW '24}
}

@article{hosseini-bilingual-2024,
	title = {A bilingual longitudinal analysis of privacy policies measuring the impacts of the {GDPR} and the {CCPA}/{CPRA}},
	volume = {2024},
    number = {2},
	doi = {10.56553/popets-2024-0058},
	journal = {Proceedings on {Privacy} {Enhancing} {Technologies}},
	author = {Hosseini, Henry and Utz, Christine and Degeling, Martin and Hupperich, Thomas},
	year = {2024},
	pages = {434--463},
}

@misc{information-commissioners-office-ico-create-2024,
	title = {Create your own privacy notice},
	url = {https://ico.org.uk/for-organisations/advice-for-small-organisations/create-your-own-privacy-notice/},
	lastaccessed = {2025-01-31},
	author = {{Information Commissioner's Office (ICO)}},
	month = aug,
	year = {2024},
}

@misc{chrome-overview-2024,
	title = {Overview of {CrUX}},
	url = {https://developer.chrome.com/docs/crux},
	language = {en},
	lastaccessed = {2025-01-23},
	journal = {Chrome for Developers},
	author = {{Chrome}},
    organization = {Google},
	month = feb,
	year = {2024},
}

@misc{swiss-confederation-federal-2020,
	title = {Federal {Act} on {Data} {Protection} ({Data} {Protection} {Act})},
	url = {https://www.fedlex.admin.ch/eli/cc/2022/491/en},
	lastaccessed = {2024-05-10},
	author = {{Swiss Confederation}},
	year = {2020},
}

@article{delalamo2022mapping,
    author    = {Jose M. Del Alamo and Danny S. Guaman and Boni Garc{\'\i}a and Ana Diez},
    title     = {A systematic mapping study on automated analysis of privacy policies},
    journal   = {Computing},
    volume    = {104},
    number    = {9},
    pages     = {2053--2076},
    year      = {2022},
    month     = may,
    doi       = {10.1007/s00607-022-01076-3},
    url       = {https://link.springer.com/article/10.1007/s00607-022-01076-3}
}

@article{oltramari2018privonto,
    author = {Oltramari, Alessandro and Piraviperumal, Dhivya and Schaub, Florian and Wilson, Shomir and Cherivirala, Sushain and Norton, Thomas B. and Russell, N. Cameron and Story, Peter and Reidenberg, Joel and Sadeh, Norman},
    title = {PrivOnto: A semantic framework for the analysis of privacy policies},
    year = {2018},
    issue-date = {2018},
    publisher = {IOS Press},
    address = {NLD},
    volume = {9},
    number = {2},
    issn = {1570-0844},
    url = {https://doi.org/10.3233/SW-170283},
    doi = {10.3233/SW-170283},
    journal = {Semantic Web},
    month = jan,
    pages = {185–203},
    numpages = {19}
}

@inproceedings{andow2019policylint,
    author    = {Benjamin Andow and Samin Yaseer Mahmud and William Wang and Justin Whitaker and William Enck and Bradley Reaves and Kapil Singh and Tao Xie},
    title     = {PolicyLint: Investigating internal privacy policy contradictions on Google Play},
    booktitle = {Proceedings of the 28th USENIX Security Symposium},
    year      = {2019},
    publisher = {USENIX Association},
    pages     = {585--602},
    address   = {Berkeley, CA, USA},
    location  = {Santa Clara, CA, United States},
    url       = {https://www.usenix.org/conference/usenixsecurity19/presentation/andow},
    series = {USENIX Security '19}
}

@inproceedings{costante2012completeness,
    author = {Costante, Elisa and Sun, Yuanhao and Petkovi\'{c}, Milan and den Hartog, Jerry},
    title = {A machine learning solution to assess privacy policy completeness: (short paper)},
    year = {2012},
    isbn = {9781450316637},
    publisher = {Association for Computing Machinery},
    address = {New York, NY, USA},
    url = {https://doi.org/10.1145/2381966.2381979},
    doi = {10.1145/2381966.2381979},
    booktitle = {Proceedings of the 2012 ACM Workshop on Privacy in the Electronic Society},
    pages = {91–96},
    numpages = {6},
    location = {Raleigh, North Carolina, USA},
    series = {WPES '12}
}

@inproceedings{sathyendra2016optout,
    author    = {Kanthashree Mysore Sathyendra and Florian Schaub and Shomir Wilson and Norman Sadeh},
    title     = {Automatic extraction of opt-out choices from privacy policies},
    publisher = {Association for the Advancement of Artificial Intelligence},
    address   = {Washington, DC, USA},
    pages     = {270--275},
    location   = {Arlington, VA, USA},
    booktitle = {AAAI 2016 Fall Symposium on Privacy and Language Technologies},
    year      = {2016},
    url       = {https://aaai.org/papers/14114-14114-automatic-extraction-of-opt-out-choices-from-privacy-policies/}
}

@inproceedings{sathyendra2017choices,
    title     = {Identifying the provision of choices in privacy policy text},
    author    = {Mysore Sathyendra, Kanthashree and Wilson, Shomir and Schaub, Florian and Zimmeck, Sebastian and Sadeh, Norman},
    editor    = {Palmer, Martha and Hwa, Rebecca and Riedel, Sebastian},
    booktitle = {Proceedings of the 2017 Conference on Empirical Methods in Natural Language Processing},
    month     = {sep},
    year      = {2017},
    location   = {Copenhagen, Denmark},
    publisher = {Association for Computational Linguistics},
    address = {Stroudsburg, PA, USA},
    url       = {https://aclanthology.org/D17-1294/},
    doi       = {10.18653/v1/D17-1294},
    pages     = {2774--2779},
    series = {EMNLP '17}
}

@inproceedings{liu2014alignment,
    author    = {Fei Liu and Rohan Ramanath and Norman M. Sadeh and Noah A. Smith},
    title     = {A step towards usable privacy policy: Automatic alignment of privacy statements},
    booktitle = {Proceedings of COLING 2014},
    year      = {2014},
    url       = {https://aclanthology.org/C14-1084.pdf},
    publisher = {Association for Computational Linguistics},
    address = {Stroudsburg, PA, USA},
    location  = {Dublin, Ireland},
    pages     = {884-894}
}

@inproceedings{massey2013policymining,
    author    = {Aaron K. Massey and Jacob Eisenstein and Annie I. Ant{\'o}n and Peter P. Swire},
    title     = {Automated text mining for requirements analysis of policy documents},
    pages     = {4--13},
    booktitle = {Proceedings of the 21st IEEE International Requirements Engineering Conference},
    year      = {2013},
    publisher = {IEEE},
    address   = {New York, NY, USA},
    url       = {https://www.cc.gatech.edu/~aianton/assets/2013-re13-nlp.pdf},
    location   = {Rio de Janeiro, Brazil},
    series = {RE '13}
}

@inproceedings{liu2016vagueness,
    author    = {Fei Liu and Nicole Lee Fella and Kexin Liao},
    title     = {Modeling language vagueness in privacy policies using deep neural networks},
    pages     = {257--263},
    publisher = {Association for the Advancement of Artificial Intelligence},
    address   = {Washington, DC, USA},
    booktitle = {AAAI 2016 Fall Symposium on Privacy and Language Technologies},
    location  = {Arlington, VA, USA},
    year      = {2016},
    url       = {https://aaai.org/papers/14059-14059-modeling-language-vagueness-in-privacy-policies-using-deep-neural-networks/}
}

@inproceedings{ravichander2019privacyqa,
    title = {Question answering for privacy policies: Combining computational and legal perspectives},
    author = {Ravichander, Abhilasha and Black, Alan W. and Wilson, Shomir and Norton, Thomas and Sadeh, Norman},
    editor = {Inui, Kentaro and Jiang, Jing and Ng, Vincent and Wan, Xiaojun},
    booktitle = {Proceedings of the 2019 Conference on Empirical Methods in Natural Language Processing and the 9th International Joint Conference on Natural Language Processing},
    month = {nov},
    year = {2019},
    location = {Hong Kong, China},
    publisher = {Association for Computational Linguistics},
    address = {Stroudsburg, PA, USA},
    url = {https://aclanthology.org/D19-1500/},
    doi = {10.18653/v1/D19-1500},
    pages = {4947--4958},
    series = {EMNLP-IJCNLP '19}
}

@inproceedings{andow2020policheck,
    author    = {Benjamin Andow and Samin Yaseer Mahmud and Justin Whitaker and William Enck and Bradley Reaves and Kapil Singh and Serge Egelman},
    title     = {Actions speak louder than words: Entity-sensitive privacy policy and data flow analysis with PoliCheck},
    booktitle = {Proceedings of the 29th USENIX Security Symposium},
    pages     = {985--1002},
    year      = {2020},
    publisher = {USENIX Association},
    address   = {Berkeley, CA, USA},
    location  = {online},
    url       = {https://www.usenix.org/system/files/sec20-andow.pdf},
    series = {USENIX Security '20}
}

@article{zimmeck2019maps,
    title   = {MAPS: Scaling privacy compliance analysis to a million apps},
    author  = {Zimmeck, Sebastian and Story, Peter and Smullen, Daniel and Ravichander, Abhilasha and Wang, Ziqi and Reidenberg, Joel R. and Russell, N. Cameron and Sadeh, Norman},
    journal = {Proceedings on Privacy Enhancing Technologies},
    volume  = {2019},
    number  = {3},
    pages   = {66--86},
    year    = {2019},
    doi     = {10.2478/popets-2019-0037},
}

@article{guaman2023crossborder,
    author    = {Danny S. Guaman and David Rodriguez and Jose M. Del Alamo and Jose Such},
    title     = {Automated GDPR compliance assessment for cross-border personal data transfers in Android applications},
    journal   = {Computers \& Security},
    volume    = {130},
    articleno = {103262},
    numpages  = {16},
    year      = {2023},
    month     = jul,
    doi       = {10.1016/j.cose.2023.103262},
    url       = {https://www.sciencedirect.com/science/article/pii/S0167404823001724}
    }

@article{rodriguez2024llmprivacy,
    author    = {David Rodriguez and Inho Yang and Jose M. Del Alamo and Norman Sadeh},
    title     = {Large language models: A new approach for privacy policy analysis at scale},
    journal   = {Computing},
    volume    = {106},
    pages     = {3879--3903},
    year      = {2024},
    doi       = {10.1007/s00607-024-01331-9},
    url       = {https://link.springer.com/article/10.1007/s00607-024-01331-9}
}

@misc{tang2023policygpt,
    author    = {Chenhao Tang and Zihan Liu and Chuxin Ma and Zihao Wu and Yuxuan Li and Weiyu Liu and Dongruo Zhu and Qizhang Li and Xiang Li and Ting Liu and Lei Fan},
    title     = {PolicyGPT: Automated analysis of privacy policies with large language models},
    year      = {2023},
    eprint    = {2309.10238},
    archivePrefix = {arXiv},
    primaryClass = {cs.CL},
    url       = {https://arxiv.org/abs/2309.10238}
}

@misc{goknil2024papel,
    author    = {Arda Goknil and Frank B. Gelderblom and Sondre Tverdal and Shreyas Tokas and Heejin Song},
    title     = {Privacy policy analysis through prompt engineering for LLMs (PAPEL)},
    year      = {2024},
    eprint    = {2409.14879},
    archivePrefix = {arXiv},
    primaryClass = {cs.CL},
    url       = {https://arxiv.org/abs/2409.14879}
}

@article{cory2025wordlevelannotationgdprtransparency,
    title={Word-level annotation of GDPR transparency compliance in privacy policies using large language models}, 
    author={Thomas Cory and Wolf Rieder and Julia Krämer and Philip Raschke and Patrick Herbke and Axel Küpper},
    year={2026},
    journal   = {Proceedings on Privacy Enhancing Technologies},
    volume    = {2026},
    number    = {1},
    pages     = {509--528},
    doi       = {10.56553/popets-2026-0026}, 
}

@inproceedings{xie2025evaluating,
    author    = {Qinge Xie and Karthik Ramakrishnan and Frank Li},
    title     = {Evaluating privacy policies under modern privacy laws at scale: An {LLM}-based automated approach},
    pages     = {5797--5816},
    booktitle = {Proceedings of the 34th USENIX Security Symposium},
    year      = {2025},
    month     = aug,
    location  = {Seattle, WA, USA},
    publisher = {USENIX Association},
    address   = {Berkeley, CA, USA},
    isbn      = {978-1-939133-52-6},
    url       = {https://www.usenix.org/conference/usenixsecurity25/presentation/xie},
    series = {USENIX Security '25}
}

@inproceedings{yang2025dbsec,
    author    = {Ming Yang and Vijayalakshmi Atluri and Shamik Sural and Atul Kundu},
    title     = {Automated privacy policy analysis using large language models},
    location   = {Gjovik, Norway},
    booktitle = {Data and Applications Security and Privacy XXXIX},
    pages     = {23--43},
    publisher = {Springer},
    address   = {Cham, Switzerland},
    year      = {2025},
    doi       = {10.1007/978-3-031-96590-6-2},
    series = {DBSec '25}
}

@inproceedings{ravichander2020policyqa,
    title     = {PolicyQA: A reading comprehension dataset for privacy policies},
    author    = {Ahmad, Wasi and Chi, Jianfeng and Tian, Yuan and Chang, Kai-Wei},
    editor    = {Cohn, Trevor and He, Yulan and Liu, Yang},
    booktitle = {Findings of the Association for Computational Linguistics},
    month     = {nov},
    year      = {2020},
    location  = {Online},
    publisher = {Association for Computational Linguistics},
    address = {Stroudsburg, PA, USA},
    pages     = {743--749},
    series    = {Findings of EMNLP '20},
    doi       = {10.18653/v1/2020.findings-emnlp.66}
}

@article{liu2023appcorp,
    author  = {Liu, Shuang and Zhang, Fan and Zhao, Baiyang and Guo, Renjie and Chen, Tao and Zhang, Meishan},
    title   = {APPCorp: A corpus for Android privacy policy document structure analysis},
    journal = {Frontiers of Computer Science},
    volume  = {17},
    number  = {3},
    pages   = {173320},
    year    = {2023},
    doi     = {10.1007/s11704-022-1627-2},
}

@inproceedings{marotta-wurgler-stein-2025-building,
    title     = {Building a long text privacy policy corpus with multi-class labels},
    author    = {Marotta-Wurgler, Florencia and Stein, David},
    booktitle = {Proceedings of the 63rd Annual Meeting of the Association for Computational Linguistics (Volume 1: Long Papers)},
    editor    = {Che, Wanxiang and Nabende, Joyce and Shutova, Ekaterina and Pilehvar, Mohammad Taher},
    month     = jul,
    year      = {2025},
    location   = {Vienna, Austria},
    publisher = {Association for Computational Linguistics},
    address = {Stroudsburg, PA, USA},
    pages     = {8156--8219},
    url       = {https://aclanthology.org/2025.acl-long.401/},
    doi       = {10.18653/v1/2025.acl-long.401},
    series = {ACL '25}
}

@inproceedings{darji2024gdprner,
    author    = {Harshil Darji and Stefan Becher and Jelena Mitrovi{\'c} and Armin Gerl and Michael Granitzer},
    title     = {A dataset of GDPR compliant NER for privacy policies},
    booktitle = {Proceedings of the 6th International Open Search Symposium},
    address   = {Garching (Munich), Germany},
    year      = {2024},
    month     = oct,
    pages     = {26--31},
    publisher = {CERN},
    doi       = {10.5281/zenodo.13871889},
    series = {ossym '24}
}

@inproceedings{story2024c3pa,
    title     = {C3PA: An open dataset of expert-annotated and regulation-aware privacy policies to enable scalable regulatory compliance audits},
    author    = {Musa, Maaz Bin and Winston, Steven M. and Allen, Garrison and Schiller, Jacob and Moore, Kevin and Quick, Sean and Melvin, Johnathan and Srinivasan, Padmini and Diamantis, Mihailis E. and Nithyanand, Rishab},
    editor    = {Al-Onaizan, Yaser and Bansal, Mohit and Chen, Yun-Nung},
    booktitle = {Proceedings of the 2024 Conference on Empirical Methods in Natural Language Processing},
    month     = {nov},
    year      = {2024},
    location  = {Miami, Florida, USA},
    publisher = {Association for Computational Linguistics},
    address = {Stroudsburg, PA, USA},
    pages     = {3710--3722},
    url       = {https://aclanthology.org/2024.emnlp-main.217},
    doi       = {10.18653/v1/2024.emnlp-main.217},
    series = {EMNLP '24}
}

@inproceedings{mashaabi2023arabic,
    author    = {Mashaabi, Malak and Al-Yahya, Ghadi and Alnashwan, Raghad and Al-Khalifa, Hend},
    title     = {Arabic privacy policy corpus and classification},
    booktitle = {Natural Language Processing and Information Systems: 28th International Conference on Applications of Natural Language to Information Systems (NLDB 2023), Derby, UK, June 21–23, 2023, Proceedings},
    year      = {2023},
    publisher = {Springer},
    address   = {Cham},
    pages     = {94--108},
    isbn      = {978-3-031-35319-2},
    doi       = {10.1007/978-3-031-35320-8-7},
    url       = {https://doi.org/10.1007/978-3-031-35320-8-7}
}

@dataset{palka2023annotated,
    author    = {Pałka, Przemysław and Pałosz, Radosław and Wiśniewska, Katarzyna},
    title     = {Annotated privacy policies of 100 online platforms},
    year      = {2023},
    publisher = {Mendeley Data},
    version   = {V1},
    doi       = {10.17632/pcgvm6zh43.1},
    url       = {https://doi.org/10.17632/pcgvm6zh43.1}
}

@inproceedings{Bernhard2025scraper,
    author = {Bernhard, David and Nenadic, Luka and Bechtold, Stefan and Kubicek, Karel},
    title = {Multilingual scraper of privacy policies and terms of service},
    year = {2025},
    isbn = {9798400714214},
    publisher = {Association for Computing Machinery},
    address = {New York, NY, USA},
    location = {Munich, Germany},
    url = {https://doi.org/10.1145/3709025.3712215},
    doi = {10.1145/3709025.3712215},
    booktitle = {Proceedings of the 2025 Symposium on Computer Science and Law},
    pages = {55–63},
    numpages = {9},
    series = {CSLAW '25}
}

@article{wagner2023privacypolicies,
    author = {Wagner, Isabel},
    title = {Privacy policies across the ages: Content of privacy policies 1996–2021},
    year = {2023},
    issue-date = {August 2023},
    publisher = {Association for Computing Machinery},
    address = {New York, NY, USA},
    volume = {26},
    number = {3},
    issn = {2471-2566},
    url = {https://doi.org/10.1145/3590152},
    doi = {10.1145/3590152},
    journal = {ACM Transactions on Privacy and Security},
    month = may,
    articleno = {32},
    numpages = {32}
}

@article{mhaidli2023researchers,
  author    = {Mhaidli, Abraham and Fidan, Selin and Doan, An and Herakovic, Gina and Srinath, Mukund and Matheson, Lee and Wilson, Shomir and Schaub, Florian},
  title     = {Researchers' experiences in analyzing privacy policies: Challenges and opportunities},
  journal   = {Proceedings on Privacy Enhancing Technologies},
  volume    = {2023},
  number    = {4},
  pages     = {287--305},
  year      = {2023},
  doi       = {10.56553/popets-2023-0111},
  url       = {https://doi.org/10.56553/popets-2023-0111},
}

@misc{tseng-stent-maida-2020-annotation-best-practices,
  author        = {Tseng, Tina and Stent, Amanda and Maida, Domenic},
  title         = {Best practices for managing data annotation projects},
  year          = {2020},
  month         = sep,
  url           = {https://arxiv.org/abs/2009.11654},
  eprint        = {2009.11654},
  archivePrefix = {arXiv},
  primaryClass  = {cs.CY},
  lastaccessed       = {2025-09-08}
}

@online{v7labs-2022-annotation-guidelines-best-practices,
  author   = {Janelidze, Ani},
  organization   = {{V7 Labs}},
  title    = {5 best practices for writing annotation guidelines},
  year     = {2022},
  month    = dec,
  day      = {19},
  url      = {https://www.v7labs.com/blog/annotation-guidelines},
  lastaccessed  = {2025-09-08}
}

@inproceedings{Zlabinger2020dexa,
    author = {Zlabinger, Markus and Sabou, Marta and Hofst\"{a}tter, Sebastian and Sertkan, Mete and Hanbury, Allan},
    title = {DEXA: Supporting non-expert annotators with dynamic examples from experts},
    year = {2020},
    isbn = {9781450380164},
    publisher = {Association for Computing Machinery},
    address = {New York, NY, USA},
    url = {https://doi.org/10.1145/3397271.3401334},
    doi = {10.1145/3397271.3401334},
    booktitle = {Proceedings of the 43rd International ACM SIGIR Conference on Research and Development in Information Retrieval},
    pages = {2109–2112},
    numpages = {4},
    location = {Virtual Event, China},
    series = {SIGIR '20}
}

@article{krippendorff2004reliability,
  author    = {Krippendorff, Klaus},
  title     = {Reliability in content analysis: Some common misconceptions and recommendations},
  journal   = {Human Communication Research},
  volume    = {30},
  number    = {3},
  pages     = {411--433},
  year      = {2004},
  publisher = {Oxford University Press},
  issn      = {0360-3989},
  doi       = {10.1111/j.1468-2958.2004.tb00738.x},
  url       = {https://doi.org/10.1111/j.1468-2958.2004.tb00738.x}
}

@article{rodriguez2024sharing,
  author    = {Rodriguez, David and Del Alamo, Jose M. and Fernández-Aller, Celia and Sadeh, Norman},
  title     = {Sharing is not always caring: Delving into personal data transfer compliance in Android apps},
  journal   = {IEEE Access},
  volume    = {12},
  pages     = {5256--5269},
  year      = {2024},
  doi       = {10.1109/ACCESS.2024.3349425},
  url       = {https://doi.org/10.1109/ACCESS.2024.3349425}
}

@misc{openai-structured-outputs-2024,
  author       = {{OpenAI}},
  title        = {Introducing structured outputs in the API},
  year         = {2024},
  month        = aug,
  day          = 6,
  publisher    = {OpenAI},
  url          = {https://openai.com/index/introducing-structured-outputs-in-the-api/},
  lastaccessed  = {2025-09-08}
}

@inproceedings{wong2021cross,
  title       = {Cross-replication reliability -- An empirical approach to inter-rater reliability},
  author      = {Wong, Ka and Paritosh, Praveen and Aroyo, Lora},
  editor      = {Zong, Chengqing and Xia, Fei and Li, Wenjie and Navigli, Roberto},
  booktitle   = {Proceedings of the 59th Annual Meeting of the Association for Computational Linguistics and the 11th International Joint Conference on Natural Language Processing (Volume 1: Long Papers)},
  month       = aug,
  year        = {2021},
  location    = {Online},
  publisher   = {Association for Computational Linguistics},
  address = {Stroudsburg, PA, USA},
  pages       = {7053--7065},
  doi         = {10.18653/v1/2021.acl-long.548},
  series = {ACL-IJCNLP '21}
}

@article{artstein2008intercoder,
  author    = {Artstein, Ron and Poesio, Massimo},
  title     = {Inter-coder agreement for computational linguistics},
  journal   = {Computational Linguistics},
  volume    = {34},
  number    = {4},
  pages     = {555--596},
  year      = {2008},
  month     = dec,
  doi       = {10.1162/coli.07-034-r2},
  url       = {https://doi.org/10.1162/coli.07-034-r2}
}

@article{Benjamini1995FDR,
  author    = {Yoav Benjamini and Yosef Hochberg},
  title     = {Controlling the false discovery rate: A practical and powerful approach to multiple testing},
  journal   = {Journal of the Royal Statistical Society: Series B (Methodological)},
  year      = {1995},
  volume    = {57},
  number    = {1},
  pages     = {289--300},
  doi       = {10.1111/j.2517-6161.1995.tb02031.x}
}

@misc{swissanwalt-datenschutz-2023,
	title = {Datenschutz {Generator} für {Webseiten}, {Blogs} und {Social} {Media} - {Kostenlos}},
	url = {https://web.archive.org/web/20230310013530/https://www.swissanwalt.ch/datenschutz-generator.aspx},
	lastaccessed = {2025-09-12},
	journal = {Internet Archive: Wayback Machine},
	author = {{SwissAnwalt}},
	month = mar,
	year = {2023},
}

@misc{privacybee-datenschutz-2025,
	title = {Datenschutz für deine {Webseite} auf {Autopilot}},
    year = {2025},
	url = {https://www.privacybee.io/ch/},
	lastaccessed = {2025-09-12},
	author = {{PrivacyBee}},
}

@misc{steiger-deutscher-2022,
	title = {Deutscher {Datenschutz}-{Generator}: 250 {Euro}-{Abmahnung} für eine {Datenschutz}­erklärung?},
	url = {https://steigerlegal.ch/2022/06/25/datenschutz-generator-abmahnungen-deutschland/},
	lastaccessed = {2025-09-12},
	organization = {Steiger Legal},
	author = {Steiger, Martin},
	month = jun,
	year = {2022},
}

@misc{schneemenschen-gmbh-datenschutzerklarung-2018,
	title = {Datenschutzerklärung},
	url = {https://perma.cc/E9W9-6U9Z},
	lastaccessed = {2025-09-12},
	author = {{Schneemenschen GmbH}},
	month = may,
	year = {2018},
}

@article{maaten-visualizing-2008,
    title = {Visualizing data using t-{SNE}},
    volume = {9},
    url = {http://jmlr.org/papers/v9/vandermaaten08a.html},
    lastaccessed = {2025-09-15},
    journal = {Journal of Machine Learning Research},
    author = {Maaten, Laurens van der and Hinton, Geoffrey},
    year = {2008},
    pages = {2579--2605},
}

@misc{openai-new-2024,
    title = {New embedding models and {API} updates},
    url = {https://openai.com/index/new-embedding-models-and-api-updates/},
    lastaccessed = {2025-09-15},
    author = {{OpenAI}},
    month = jan,
    year = {2024},
}

@article{betts-dawn-2017,
    title = {The dawn of fully automated contract drafting: {Machine} learning breathes new life into a decades-old promise},
    volume = {15},
    url = {https://scholarship.law.duke.edu/dltr/vol15/iss1/11},
    lastaccessed = {2025-09-16},
    journal = {Duke Law \& Technology Review},
    author = {Betts, Kathryn D. and Jaep, Kyle R.},
    year = {2017},
    pages = {216--233},
}

@article{barton-access-2019,
    title = {Access to justice and routine legal services: {New} technologies meet bar regulators},
    volume = {70},
    shorttitle = {Access to {justice} and {routine} {legal} {services}},
    url = {https://repository.uclawsf.edu/hastings_law_journal/vol70/iss4/2/},
    lastaccessed = {2025-09-16},
    journal = {Hastings Law Journal},
    author = {Barton, Benjamin and Rhode, Deborah},
    year = {2019},
    pages = {955--989},
}

@misc{contreras-solving-2025,
    title = {Solving for agreement},
    url = {https://ssrn.com/abstract=5389732},
    author = {Contreras, Jorge L.},
    year = {2025},
    archivePrefix = {{SSRN} {Working} {Paper}},
    eprint  = {5389732}
}

@inproceedings{guha2023legalbench,
    author = {Guha, Neel and Nyarko, Julian and Ho, Daniel E. and R\'{e}, Christopher and Chilton, Adam and Narayana, Aditya and Chohlas-Wood, Alex and Peters, Austin and Waldon, Brandon and Rockmore, Daniel N. and Zambrano, Diego and Talisman, Dmitry and Hoque, Enam and Surani, Faiz and Fagan, Frank and Sarfaty, Galit and Dickinson, Gregory M. and Porat, Haggai and Hegland, Jason and Wu, Jessica and Nudell, Joe and Niklaus, Joel and Nay, John and Choi, Jonathan H. and Tobia, Kevin and Hagan, Margaret and Ma, Megan and Livermore, Michael and Rasumov-Rahe, Nikon and Holzenberger, Nils and Kolt, Noam and Henderson, Peter and Rehaag, Sean and Goel, Sharad and Gao, Shang and Williams, Spencer and Gandhi, Sunny and Zur, Tom and Iyer, Varun and Li, Zehua},
    title = {LEGALBENCH: A collaboratively built benchmark for measuring legal reasoning in large language models},
    year = {2023},
    publisher = {Curran Associates Inc.},
    address = {Red Hook, NY, USA},
    booktitle = {Proceedings of the 37th International Conference on Neural Information Processing Systems},
    articleno = {1915},
    numpages = {157},
    location = {New Orleans, LA, USA},
    series = {NeurIPS '23}
}

@misc{microsoft-presidio-nodate,
    title = {Presidio: {Data} protection and de-identification {SDK}},
    url = {https://microsoft.github.io/presidio/},
    lastaccessed = {2025-09-26},
    author = {{Microsoft}},
    year = {2025},
}

@misc{siebert-dsgvo-konforme-2025,
    title = {{DSGVO}-konforme {Datenschutzerklärung} jetzt kostenlos erstellen},
    url = {https://www.e-recht24.de/muster-datenschutzerklaerung.html},
    lastaccessed = {2025-09-26},
    organization = {eRecht24},
    author = {Siebert, Sören},
    month = jul,
    year = {2025},
}

@misc{pandit-data-2024,
    title = {Data {Privacy} {Vocabulary} ({DPV}) -- {Version} 2},
    url = {https://arxiv.org/abs/2404.13426},
    author = {Pandit, Harshvardhan J. and Esteves, Beatriz and Krog, Georg P. and Ryan, Paul and Golpayegani, Delaram and Flake, Julian},
    year = {2024},
    archivePrefix = {arXiv},
    eprint = {2404.13426},
    primaryClass = {cs.CY},
}

@misc{google-compact-2022,
    title = {Compact {Language} {Detector} v3 ({CLD3})},
    url = {https://github.com/google/cld3},
    lastaccessed = {2025-09-11},
    author = {{Google}},
    year = {2022},
}

@article{foster_when_2018,
    title = {When to praise the machine: {The} promise and perils of automated transactional drafting},
    volume = {69},
    url = {https://scholarcommons.sc.edu/sclr/vol69/iss3/5},
    number = {3},
    lastaccessed = {2026-01-27},
    journal = {South Carolina Law Review},
    author = {Foster, William and Lawson, Andrew},
    month = apr,
    year = {2018},
    pages = {597--635},
}

@inproceedings{sun_quality_2020,
    address = {New York, NY, USA},
    location = {Trondheim, Norway},
    series = {{EASE} '20},
    title = {Quality assessment of online automated privacy policy generators: {An} empirical study},
    doi = {10.1145/3383219.3383247},
    booktitle = {Proceedings of the 24th {International} {Conference} on {Evaluation} and {Assessment} in {Software} {Engineering}},
    publisher = {Association for Computing Machinery},
    author = {Sun, Ruoxi and Xue, Minhui},
    month = apr,
    year = {2020},
    pages = {270--275},
}

@inproceedings{pan_is_2024,
    location = {Philadelphia, PA},
    series = {USENIX Security '24},
    title = {Is it a trap? {A} large-scale empirical study and comprehensive assessment of online automated privacy policy generators for mobile apps},
    doi = {10.5555/3698900.3699218},
    booktitle = {Proceedings of the 33rd {USENIX} {Conference} on {Security} {Symposium}},
    publisher = {USENIX Association},
    address   = {Berkeley, CA, USA},
    author = {Pan, Shidong and Zhang, Dawen and Staples, Mark and Xing, Zhenchang and Chen, Jieshan and Xu, Xiwei and Hoang, Thong},
    year = {2024},
    pages = {5681--5698},
}

@article{saxon_computer-aided_1982,
    title = {Computer-aided drafting of legal documents},
    volume = {7},
    doi = {10.1111/j.1747-4469.1982.tb00469.x},
    number = {3},
    journal = {American Bar Foundation Research Journal},
    author = {Saxon, Charles S.},
    month = jul,
    year = {1982},
    pages = {685--754},
}

@article{AiNorton2024DiDReview,
  author  = {Ai, Chunrong and Norton, Edward C.},
  title   = {Difference in differences, ratio in ratios, and ratio in odds ratios for limited dependent variables: A review and more},
  journal = {Studies in Nonlinear Dynamics \& Econometrics},
  year    = {2024},
  month   = aug,
  volume  = {28},
  number  = {4},
  pages   = {1--32},
  doi     = {10.1515/snde-2024-0125}
}

@inproceedings{levy_same_2024,
    location = {Bangkok, Thailand},
    title = {Same task, more tokens: {The} impact of input length on the reasoning performance of large language models},
    shorttitle = {Same {Task}, {More} {Tokens}},
    url = {https://aclanthology.org/2024.acl-long.818},
    doi = {10.18653/v1/2024.acl-long.818},
    language = {en},
    lastaccessed = {2026-02-16},
    booktitle = {Proceedings of the 62nd {Annual} {Meeting} of the {Association} for {Computational} {Linguistics} ({Volume} 1: {Long} {Papers})},
    publisher = {Association for Computational Linguistics},
    address = {Stroudsburg, PA, USA},
    author = {Levy, Mosh and Jacoby, Alon and Goldberg, Yoav},
    year = {2024},
    series = {ACL '24},
    pages = {15339--15353},
}

@misc{datenschutzpartner_datenschutz-generator_2026,
    title = {Datenschutz-{Generator} für {Datenschutzerklärungen}},
    url = {https://www.datenschutzpartner.ch/angebot-datenschutz-generator/},
    lastaccessed = {2026-02-20},
    author = {{Datenschutzpartner}},
    year = {2026},
}

@misc{apacible-bernardo_data_2025,
    title = {Data protection and privacy laws now in effect in 144 countries},
    url = {https://iapp.org/news/a/data-protection-and-privacy-laws-now-in-effect-in-144-countries},
    lastaccessed = {2026-02-23},
    journal = {IAPP},
    author = {Apacible-Bernardo, Aly and Bushey, Kayla},
    organization = {International Association of Privacy Professionals (IAPP)},
    month = jan,
    year = {2025},
}

@inproceedings{tran_measuring_2024,
    address = {New York, NY},
    location = {Honolulu, HI, USA},
    title = {Measuring compliance with the {California} {Consumer} {Privacy} {Act} over space and time},
    doi = {10.1145/3613904.3642597},
    booktitle = {Proceedings of the 2024 {CHI} {Conference} on {Human} {Factors} in {Computing} {Systems}},
    publisher = {Association for Computing Machinery},
    author = {Tran, Van Hong and Mehrotra, Aarushi and Chetty, Marshini and Feamster, Nick and Frankenreiter, Jens and Strahilevitz, Lior},
    month = may,
    year = {2024},
    pages = {1--19},
    series = {CHI '24}
}

@inproceedings{zimmeck_automated_2017,
    address = {San Diego, CA},
    title = {Automated analysis of privacy requirements for mobile apps},
    url = {https://www.ndss-symposium.org/ndss2017/ndss-2017-programme/automated-analysis-privacy-requirements-mobile-apps/},
    doi = {10.14722/ndss.2017.23034},
    booktitle = {Proceedings 2017 {Network} and {Distributed} {System} {Security} {Symposium}},
    publisher = {Internet Society},
    author = {Zimmeck, Sebastian and Wang, Ziqi and Zou, Lieyong and Iyengar, Roger and Liu, Bin and Schaub, Florian and Wilson, Shomir and Sadeh, Norman and Bellovin, Steven M. and Reidenberg, Joel},
    year = {2017},
    pages = {1--15},
}

@misc{federal_data_protection_and_information_commissioner_data_2025,
    title = {Data {Protection} {Day} 2025: {Data} protection in the face of digital transformation},
    shorttitle = {Data {Protection} {Day} 2025},
    url = {https://www.edoeb.admin.ch/en/data-protection-day-2025},
    language = {en},
    urldate = {2026-05-08},
    author = {{Federal Data Protection and Information Commissioner}},
    month = jan,
    year = {2025},
}

@techreport{federal_office_of_justice_rechtliche_2024,
    title = {Rechtliche {Basisanalyse} im {Rahmen} der {Auslegeordnung} zu den {Regulierungsansätzen} im {Bereich} künstliche {Intelligenz}},
    url = {https://www.bj.admin.ch/bj/de/home/staat/gesetzgebung/kuenstliche-intelligenz.html},
    language = {de},
    urldate = {2026-05-08},
    author = {{Federal Office of Justice}},
    month = aug,
    year = {2024},
}

@article{thouvenin_rechtsrahmen_2026,
    title = {Ein {Rechtsrahmen} für {KI} in der {Schweiz}: {Handlungsbedarf} und {Handlungsoptionen} bei {Transparenz} und {Datenschutz}},
    volume = {March 30, 2026},
    url = {https://jusletter.weblaw.ch/juslissues/2026/1278/ein-rechtsrahmen-fur_995089b415.html__ONCE&login=false},
    urldate = {2026-05-08},
    journal = {Jusletter},
    author = {Thouvenin, Florent},
    year = {2026},
    pages = {1--21},
}


\appendix
\section{Appendix}

\subsection{Scraper Performance} \label{app:scraper}

We report the detailed scraper performance metrics as drawn and adapted from Bernhard et al.~\cite[p. 59]{Bernhard2025scraper} in Table~\ref{tab:classification-metrics}.

\begin{table}[ht]
\begin{tabular}{|c|c|c|c|c|c|}
\hline
\rowcolor[HTML]{C0C0C0} 
             &   English &   German &   French &   Italian & \textbf{  Average} \\ \hline
  Accuracy     & \cellcolor[HTML]{E0F295} 0.73 & \cellcolor[HTML]{ECF7A6} 0.70  & \cellcolor[HTML]{D9EF8B} 0.75 & \cellcolor[HTML]{E5F49B} 0.72  & \cellcolor[HTML]{E0F295} 0.73 \\ \hline
  Recall       & \cellcolor[HTML]{93D168} 0.88 & \cellcolor[HTML]{C9E881} 0.78 & \cellcolor[HTML]{BBE278} 0.81 & \cellcolor[HTML]{7FC866} 0.91  & \cellcolor[HTML]{A7DC6D} 0.85 \\ \hline
  Precision    & \cellcolor[HTML]{C9E881} 0.78 & \cellcolor[HTML]{A0D669} 0.86 & \cellcolor[HTML]{CFEB85} 0.77 & \cellcolor[HTML]{E0F295} 0.73  & \cellcolor[HTML]{C5E67E} 0.79 \\ \hline
  F$_1$ score     & \cellcolor[HTML]{AFDD70} 0.83 & \cellcolor[HTML]{BBE278} 0.81 & \cellcolor[HTML]{C5E67E} 0.79 & \cellcolor[HTML]{BBE278} 0.81  & \cellcolor[HTML]{BBE278} 0.81 \\ \hline
\end{tabular}
\caption{Scraper's policy classification metrics.}
\label{tab:classification-metrics}
\end{table}

\subsection{Bootstrap Confidence Intervals}
\label{app:bootstrap-ci}

\begin{table}[H]
\small
\centering
\setlength{\tabcolsep}{4pt}
\begin{threeparttable}
\begin{tabular}{llrrrr}
\toprule
\textbf{Language} & \textbf{Practice} & \textbf{F$_1$} & \textbf{CI Low} & \textbf{CI High} & \textbf{Positives (GT)} \\
\midrule
English & \texttt{ispol} & 0.933 & 0.837 & 1.000 & 22 \\
English & \texttt{contr} & 0.929 & 0.812 & 1.000 & 13 \\
English & \texttt{purp} & 0.933 & 0.842 & 1.000 & 22 \\
English & \texttt{rect} & 0.971 & 0.903 & 1.000 & 17 \\
English & \texttt{forg} & 1.000 & 1.000 & 1.000 & 17 \\
English & \texttt{port} & 0.909 & 0.727 & 1.000 & 11 \\
English & \texttt{comp} & 1.000 & 1.000 & 1.000 & 9 \\
English & \texttt{hum} & 0.600 & 0.000 & 0.889 & 3 \\
German & \texttt{ispol} & 0.983 & 0.947 & 1.000 & 29 \\
German & \texttt{contr} & 0.909 & 0.786 & 1.000 & 15 \\
German & \texttt{purp} & 0.983 & 0.947 & 1.000 & 29 \\
German & \texttt{rect} & 0.950 & 0.867 & 1.000 & 19 \\
German & \texttt{forg} & 0.977 & 0.917 & 1.000 & 21 \\
German & \texttt{port} & 0.966 & 0.880 & 1.000 & 14 \\
German & \texttt{comp} & 1.000 & 1.000 & 1.000 & 12 \\
German & \texttt{hum} & 0.727 & 0.286 & 1.000 & 4 \\
French & \texttt{ispol} & 1.000 & 1.000 & 1.000 & 27 \\
French & \texttt{contr} & 0.833 & 0.686 & 0.941 & 15 \\
French & \texttt{purp} & 1.000 & 1.000 & 1.000 & 27 \\
French & \texttt{rect} & 0.927 & 0.827 & 1.000 & 19 \\
French & \texttt{forg} & 1.000 & 1.000 & 1.000 & 20 \\
French & \texttt{port} & 0.941 & 0.769 & 1.000 & 8 \\
French & \texttt{comp} & 1.000 & 1.000 & 1.000 & 10 \\
French & \texttt{hum} & 0.727 & 0.333 & 1.000 & 4 \\
Italian & \texttt{ispol} & 1.000 & 1.000 & 1.000 & 22 \\
Italian & \texttt{contr} & 0.950 & 0.869 & 1.000 & 20 \\
Italian & \texttt{purp} & 1.000 & 1.000 & 1.000 & 22 \\
Italian & \texttt{rect} & 0.977 & 0.919 & 1.000 & 21 \\
Italian & \texttt{forg} & 1.000 & 1.000 & 1.000 & 21 \\
Italian & \texttt{port} & 0.933 & 0.815 & 1.000 & 15 \\
Italian & \texttt{comp} & 1.000 & 1.000 & 1.000 & 14 \\
Italian & \texttt{hum} & 0.333 & 0.000 & 0.800 & 4 \\
Overall & \texttt{ispol} & 0.980 & 0.957 & 0.995 & 100 \\
Overall & \texttt{contr} & 0.905 & 0.849 & 0.951 & 63 \\
Overall & \texttt{purp} & 0.980 & 0.959 & 0.995 & 100 \\
Overall & \texttt{rect} & 0.956 & 0.918 & 0.983 & 76 \\
Overall & \texttt{forg} & 0.994 & 0.980 & 1.000 & 79 \\
Overall & \texttt{port} & 0.939 & 0.878 & 0.981 & 48 \\
Overall & \texttt{comp} & 1.000 & 1.000 & 1.000 & 45 \\
Overall & \texttt{hum} & 0.632 & 0.438 & 0.788 & 15 \\
\bottomrule
\end{tabular}
\begin{tablenotes}
\item \textit{Notes.} 95\% confidence intervals (CI) are computed via policy-level bootstrap with replacement (1,000 resamples; fixed seed). ``Positives (GT)'' reports the number of positive cases per practice and language.
\end{tablenotes}
\end{threeparttable}
\caption{Validation performance (F$_1$) with 95\% bootstrap confidence intervals by language and practice.}
\label{tab:bootstrap-ci-f1}
\end{table}

To quantify the statistical stability of our validation results, we compute non-parametric 95\% confidence intervals for all F$_1$ scores using policy-level bootstrap resampling (1,000 iterations; fixed random seed). Resampling is performed at the document level to preserve within-policy dependence across disclosure dimensions.

Table~\ref{tab:bootstrap-ci-f1} reports point estimates together with bootstrap intervals and the number of positive cases annotated per obligation and language. For most disclosure dimensions, the intervals remain narrow across languages, indicating stable performance estimates. Wider intervals arise for \texttt{hum}, which reflects the conditional nature of this obligation and its lower prevalence in the benchmark.

Overall, the bootstrap analysis corroborates the robustness of the multilingual validation results while transparently characterizing residual uncertainty across obligations.

\subsection{Last Updated Date (October Policies)}\label{app:last_updated}

\begin{figure}[H]
    \centering
    \includegraphics[width=1\linewidth]{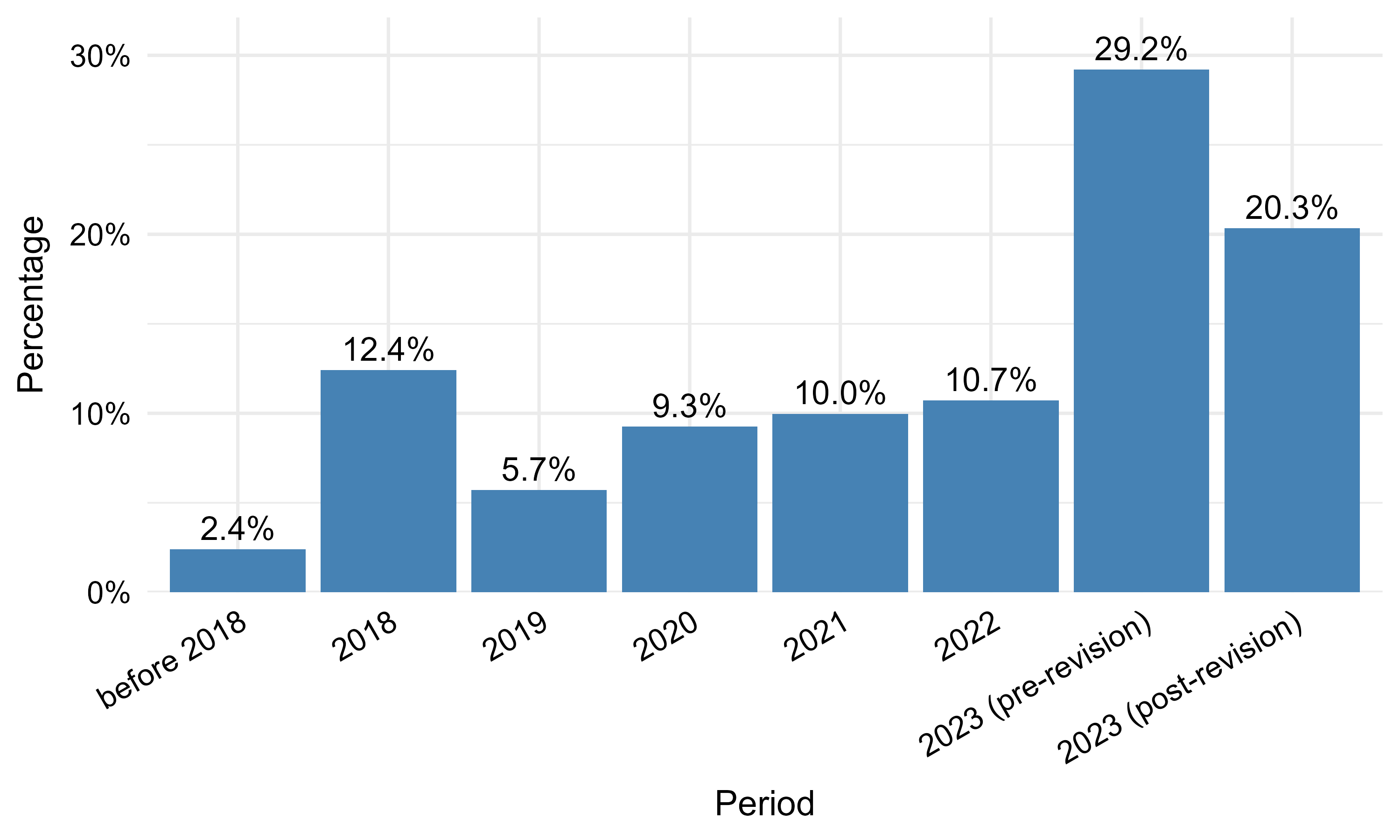}
    \caption{Last updated dates as a percentage of all October policies that are explicitly dated ($N = 5,867$).}
    \Description{Bar plot of the last updated dates (before 2018, 2018, 2019, 2020, 2021, 2022, 2023 (pre-revision), and 2023 (post-revision)) for October policies as a relative share of all policies explicitly stating such date.}
    \label{fig:last_updated}
\end{figure}

Looking at the last updated dates (\texttt{upd}) for October policies, Figure~\ref{fig:last_updated} shows that most policies with an explicitly mentioned date ($N = 5,867$) were last updated in 2023: either before the Swiss privacy law revision (29.2\%) or shortly thereafter (20.3\%).

\subsection{Policy Language Distribution}\label{app:languages}

We present the full group-level language distributions in Table~\ref{tab:languages}.

\begin{table}[ht]
\centering
\begin{tabular}{lllllll}
\toprule
\multicolumn{1}{c}{ } & \multicolumn{2}{c}{EU} & \multicolumn{2}{c}{CH} & \multicolumn{2}{c}{CH \& EU} \\
\cmidrule(l{3pt}r{3pt}){2-3}
\cmidrule(l{3pt}r{3pt}){4-5}
\cmidrule(l{3pt}r{3pt}){6-7}
 & Aug & Oct & Aug & Oct & Aug & Oct \\
\midrule
Websites        & 5903 & 5903 & 20477 & 20477 & 9093 & 9093 \\
\% With policy  & 35.25 & 36.86 & 34.19 & 40.02 & 37.12 & 43.57 \\
\quad\% in German       & 57.95 & 56.94 & 90.55 & 90.73 & 79.11 & 80.29 \\
\quad\% in English      & 17.64 & 18.75 & 7.34  & 7.08  & 16.41 & 15.45 \\
\quad\% in French       & 1.44  & 1.10  & 0.80  & 0.90  & 1.45  & 1.39 \\
\quad\% in Italian      & 22.97 & 23.21 & 1.31  & 1.29  & 3.02  & 2.88 \\
\bottomrule
\end{tabular}
\caption{Dataset summary statistics.}
\label{tab:languages}
\end{table}

\subsection{Policy Length and Readability Metrics} \label{app:policy_length}

Table~\ref{tab:policy-length} shows the mean and median word counts for all groups.

\begin{table}[ht]
\centering
\begin{tabular}{lrrr}
\toprule
 & EU & CH & CH \& EU \\
\midrule
\# Words August   & 2243 & 1818 & 2341 \\
\# Words October  & 2257 & 2125 & 2643 \\
\midrule
\textbf{$\Delta$ \# words}      & +14 & +307 & +302 \\
\textbf{$\Delta$ percentage}   & +0.6\% & +16.9\% & +12.9\% \\
\bottomrule
\end{tabular}
\caption{Median policy word counts.}
\label{tab:policy-length}
\end{table}

\subsection{GDPR and FADP Terms}\label{app:laws-terms}

We compiled two multilingual lists of terms associated with the GDPR (Table~\ref{tab:mentions-gdpr}) and the FADP (Table~\ref{tab:mentions-fadp}) to create Table~\ref{tab:mentions-gdpr-fadp}.

\begin{table}[H]
\centering
\begin{tabular}{l}
\toprule
\textbf{GDPR Terms} \\
\midrule
GDPR \\
2016/679 \\
General Data Protection Regulation \\
DSGVO \\
DS-GVO \\
Datenschutz-Grundverordnung \\
Datenschutzgrundverordnung \\
Datenschutz Grundverordnung \\
règlement général sur la protection des données \\
RGPD \\
regolamento generale sulla protezione dei dati \\
\bottomrule
\end{tabular}
\caption{GDPR mention terms.}
\label{tab:mentions-gdpr}
\end{table}

\begin{table}[H]
\centering
\begin{tabular}{l}
\toprule
\textbf{FADP Terms} \\
\midrule
235.1 \\
Federal Act on Data Protection \\
Data Protection Act \\
DSG \\
nDSG \\
rDSG \\
Bundesgesetz über den Datenschutz \\
Datenschutzgesetz \\
LPD \\
Loi fédérale sur la protection des données \\
Legge federale sulla protezione dei dati \\
\bottomrule
\end{tabular}
\caption{FADP mention terms.}
\label{tab:mentions-fadp}
\end{table}

\subsection{Disclosure Heterogeneity} \label{app:heterogeneity}

\begin{table*}[!t]
\setlength{\tabcolsep}{5pt} 
\centering
\begin{tabular}{lccccccccccc}
\toprule
 & \multicolumn{3}{c}{October Update} & \multicolumn{3}{c}{Top 5k Website} & \multicolumn{4}{c}{Language} \\
\cmidrule(lr){2-4} \cmidrule(lr){5-7} \cmidrule(lr){8-11}
Obligation & no & yes & $\Delta$ p.p.\textsuperscript{1}\ & no & yes & $\Delta$ p.p.\textsuperscript{1} & de & en & it & fr \\
\midrule
\textit{contr} & 75.0\% & 88.8\% & +13.9 & 79.5\% & 86.8\% & +7.3 & 79.7\% & 79.9\% & 88.7\% & 75.2\% \\
\textit{purp}  & 98.8\% & 98.8\% & $-0.0$ & 98.9\% & 98.1\% & $-0.8$ & 98.9\% & 98.0\% & 98.5\% & 99.3\% \\
rect  & 77.2\% & 85.6\% & +8.4 & 79.9\% & 85.1\% & +5.2 & 81.2\% & 75.1\% & 80.1\% & 64.1\% \\
forg  & 79.5\% & 86.6\% & +7.1 & 81.8\% & 86.7\% & +5.0 & 82.9\% & 77.9\% & 81.7\% & 68.6\% \\
port  & 53.2\% & 65.9\% & +12.7 & 57.2\% & 66.2\% & +9.0 & 58.4\% & 54.9\% & 60.4\% & 32.0\% \\
comp  & 51.7\% & 62.8\% & +11.1 & 54.8\% & 67.6\% & +12.8 & 56.7\% & 49.4\% & 60.6\% & 26.8\% \\
\textit{hum}   & 19.3\% & 25.4\% & +6.2 & 20.5\% & 33.8\% & +13.2 & 21.2\% & 24.9\% & 23.6\% & 7.2\% \\
\midrule
\textbf{Average} & 64.9\% & 73.4\% & +8.5 & 67.5\% & 74.9\% & +7.4 & 68.4\% & 65.7\% & 70.5\% & 53.3\% \\
\textbf{Total policies} & 9044 & 5289 & -- & 13225 & 1108 & -- & 11855 & 1600 & 725 & 153 \\
\bottomrule
\end{tabular}
\medskip

\footnotesize{1. The $\Delta$ columns report percentage-point differences for the \emph{yes} subsets versus the \emph{no} subsets.}
\vspace{5pt}
\caption{Disclosure per obligation across updated status, website rank, and languages (October 2023).}
\label{tab:disclosure-details}
\end{table*}

\begin{table}[ht]
\centering
\setlength{\tabcolsep}{5pt}
\begin{tabular}{lcc}
\toprule
October policy status & Average disclosure rate & Policies \\
\midrule
Unchanged text & 64.9\% & 9{,}044 \\
Newly valid in October & 70.4\% & 2{,}419 \\
Changed text & 76.0\% & 2{,}870 \\
\bottomrule
\end{tabular}
\caption{Average disclosure rates by October policy status.}
\label{tab:oct-policy-status}
\end{table}

Table~\ref{tab:disclosure-details} disaggregates October disclosure rates by update status (between the two snapshots), website popularity, and language across all groups. The table provides the obligation-level results underlying the summary reported in Section~\ref{sec:disclosure_practices}. Policies that were either newly valid in October or textually changed between August and October display substantially higher disclosure rates than unchanged policies (+8.5 p.p.). Table~\ref{tab:oct-policy-status} further separates this pattern into policies whose text changed between snapshots (``Changed text''), policies that became valid only in October (``Newly valid in October''), and unchanged policies (``Unchanged text''). This decomposition is consistent with active revision during the observation window, though it may also reflect broader standardization and learning dynamics~\cite{kahan-standardization-1997} in policy drafting. The descriptive decomposition, however, does not allow these mechanisms to be disentangled.

Website popularity is also associated with higher disclosure rates. Websites included in any country-specific Top~5k bucket exhibit, on average, +7.4 p.p. higher disclosure rates than less visited websites (Table~\ref{tab:disclosure-details}). Language-related variation is comparatively modest. Differences between German- and English-language policies remain below 3 p.p. on average. Italian-language policies exhibit comparatively high disclosure levels, and French-language policies display lower overall rates. The smaller sample sizes for Italian ($N = 725$) and French ($N = 153$), however, warrant cautious interpretation.

Finally, Table~\ref{tab:heterogeneity-interactions} reports interaction models using an aggregate disclosure index to assess whether the differential August--October increases are concentrated among particular subsets of websites. The results indicate that website popularity is primarily associated with disclosure levels rather than with the reform-related increase itself. By contrast, the increase is strongest among German-language policies, while the non-German interaction is negative, particularly for the \texttt{CH} group.

\subsection{Manual Annotation Corrections} \label{app:annotations_comparison}

In Table~\ref{tab:annotations_comparison}, we present all changes performed to the annotations by our legal expert, including the relevant snippet for the change. Naturally, the legal expert could not identify the relevant snippet if they did not detect the relevant disclosure (see Policy~93). Policy~77 (denoted with a star *) represents a special case, as it constitutes a website imprint that was erroneously annotated as a valid policy, even though it did not mention personal data at all.

\begin{table*}[ht]
\centering
\begin{tabular}{|l|c|c|p{9cm}|}
\hline
\textbf{Entry} & \textbf{Original} & \textbf{Correction} & \textbf{Relevant Snippet} \\
\hline
Policy 2: hum & 0 & 1 & Art. 22 GDPR: In cases of solely automated processing with legal effects, special rights are granted for data subjects. \\ \hline
Policy 5: upd & 24/09/2024 & 24/09/2018 & Version: \#2018-09-24 \\ \hline
Policy 6: purp & 0 & 1 & Personal information submitted to us through our website will be used for the purposes specified in this policy. [...] \\ \hline
Policy 21: purp & 0 & 1 & The personal data entered by the data subject is collected and stored solely for use at the location of the data controller and its own purposes. \\ \hline
Policy 23: rect & 0 & 1 & You may have
further statutory rights, such as the right to have data erased or corrected, to have data processing restricted and to have data transmitted. \\ \hline
Policy 67: port & 0 & 1 & A certaines conditions, vous pouvez nous demander de vous transmettre à vous ou à un tiers désigné par vous, vos données personnelles dans un format courant. \\ \hline
Policy 72: contr & 0 & 1 & Identité du responsable de traitement : Les données personnelles sont collectées par [...] \\ \hline
Policy 77: contr & 1 & 0 & NA* \\ \hline
Policy 77: purp & 1 & 0 & NA* \\ \hline
Policy 77: rect & 1 & 0 & NA* \\ \hline
Policy 77: forg & 1 & 0 & NA* \\ \hline
Policy 77: port & 1 & 0 & NA* \\ \hline
Policy 77: comp & 1 & 0 & NA* \\ \hline
Policy 77: hum & 1 & 0 & NA* \\ \hline
Policy 81: contr & 0 & 1 & Pour toute question ou demande liée au traitement de vos données personnelles, vous pouvez nous contacter par e-mail à <redacted email> ou par courrier postal à l’adresse : [...]
 \\ \hline
Policy 93: contr & 1 & 0 & NA \\ \hline
Policy 97: purp & 0 & 1 & In particolare, la
scuola, ove pubblichi foto di alunni e insegnanti, tratterà delle immagini e lo farà al solo fine di documentare attività didattiche di particolare pregio avendo cura di non rendere identificabili gli alunni in mancanza di base giuridica che non si identifichi nel perseguimento di un interesse pubblico. \\ \hline
Policy 101: contr & 0 & 1 & Responsabile per la protezione dei dati: [...] \\ \hline
Policy 112: upd & NA & 25/05/2018 & Questa policy, in vigore dal 25 maggio 2018 [...] \\ \hline
Policy 113: upd & NA & 25/05/2018 & La presente Privacy Policy è aggiornata alla data del 25 maggio 2018 \\ \hline
Policy 114: upd & 01/07/2019 & 01/06/2019 & Ultimo aggiornamento: giugno 2019 \\ \hline
\end{tabular}
\vspace{12pt}

\caption{Overview of the detected errors between the originally annotated and the corrected benchmark.}

\label{tab:annotations_comparison}
\end{table*}

\subsection{Impact of Annotation Corrections on Model Performance}
\label{app:corrections_impact}

To ensure transparency regarding the post-annotation corrections described in Appendix~\ref{app:annotations_comparison}, we additionally evaluated GPT-5 on the original annotations prior to the corrections performed by the legal expert. The original annotations are publicly available in our artifact repository.

Overall, performance differences between the original and corrected annotations are small. Across all language–practice combinations, the mean change in F$_1$ is 0.021, with a median of 0.000. Table~\ref{tab:delta_f1_corrections} reports the differences in F$_1$ scores between the corrected and original annotations across languages and practices. The largest improvements occur in dimensions with very low numbers of positive cases, particularly \texttt{hum}, where even a single annotation change can substantially affect the resulting F$_1$ score. For the remaining dimensions, differences are minor and do not alter the substantive conclusions reported in Section~\ref{sec:method_validation}. No systematic performance regressions are observed across languages or practices.

These findings indicate that the corrections primarily improved annotation accuracy without materially affecting the evaluation outcomes. Importantly, the corrections were identified through manual review of disagreement cases between model outputs and human annotations, and were applied only where the policy text unambiguously supported one interpretation. The robustness of model performance across both annotation versions supports the validity of the reported results.

\begin{table}[H]
\centering
\small
\setlength{\tabcolsep}{3pt}
\begin{tabular}{lcccccccc}
\hline
\textbf{Lang.} & \texttt{comp} & \texttt{contr} & \texttt{forg} & \texttt{hum} & \texttt{ispol} & \texttt{port} & \texttt{purp} & \texttt{rect} \\
\hline
EN & 0.000 & 0.000 & 0.000 & 0.156 & 0.000 & 0.000 & 0.050 & 0.030 \\
FR & 0.048 & 0.091 & 0.024 & 0.061 & 0.000 & 0.118 & 0.018 & 0.022 \\
DE & 0.000 & 0.000 & 0.000 & 0.000 & 0.000 & 0.000 & 0.000 & 0.000 \\
IT & 0.000 & 0.050 & 0.000 & 0.000 & 0.000 & 0.000 & 0.023 & 0.000 \\
\hline
\textbf{Overall} & 0.011 & 0.038 & 0.006 & 0.053 & 0.000 & 0.020 & 0.020 & 0.013 \\
\hline
\end{tabular}
\caption{Differences in F$_1$ scores ($\Delta$F$_1$) between evaluations using the corrected and original annotations. Positive values indicate improved performance after correction.}
\label{tab:delta_f1_corrections}
\end{table}

\begin{table*}[ht]
\centering
\setlength{\tabcolsep}{5pt}
\begin{tabular}{llcc}
\toprule
Moderator & Term & Coef. (p.p.) & 95\% CI \\
\midrule
\multirow{4}{*}{Top 5k website}
& \texttt{post $\times$ CH} & +2.41 & [2.07, 2.75] \\
& \texttt{post $\times$ CH \& EU} & +2.32 & [1.80, 2.85] \\
& \texttt{post $\times$ CH $\times$ Top 5k} & -2.46 & [-5.05, 0.13] \\
& \texttt{post $\times$ CH \& EU $\times$ Top 5k} & -0.71 & [-2.20, 0.78] \\
\midrule
\multirow{4}{*}{Non-German policy}
& \texttt{post $\times$ CH} & +2.51 & [2.13, 2.90] \\
& \texttt{post $\times$ CH \& EU} & +2.40 & [1.84, 2.97] \\
& \texttt{post $\times$ CH $\times$ Non-German} & -2.18 & [-2.98, -1.39] \\
& \texttt{post $\times$ CH \& EU $\times$ Non-German} & -0.90 & [-2.01, 0.20] \\
\bottomrule
\end{tabular}
\medskip

\footnotesize{
OLS linear probability models estimated on the balanced panel of $11{,}800$ websites ($23{,}600$ policy--snapshot observations). Standard errors are clustered at the website level.
}
\vspace{5pt}
\caption{Heterogeneity interaction models using the aggregate disclosure index.}
\label{tab:heterogeneity-interactions}
\end{table*}

\subsection{Logistic Difference-in-Differences Specification}
\label{app:logit_did}

To assess the sensitivity of the results to functional form, we re-estimate the difference-in-differences models using logistic regression. This nonlinear specification models disclosure on the log-odds scale and constrains fitted values to the unit interval.

All specifications match the analysis in Section~\ref{sec:results}. For each disclosure obligation, we estimate a logistic regression on the same balanced panel of $N = 11{,}800$ websites observed in both August and October 2023 ($23{,}600$ policy--snapshot observations). The dependent variable indicates whether the obligation is disclosed. The model includes indicators for time (October vs.\ August), group membership (\texttt{EU} as the reference), and their interaction terms. As in the linear models, we include CrUX rank categories and policy language as controls and cluster standard errors at the website level.

\begin{table}[H]
\centering
\setlength{\tabcolsep}{4pt}
\begin{tabular}{lcc}
\toprule
Obligation & CH: OR (95\% CI) & CH \& EU: OR (95\% CI) \\
\midrule
\textit{contr} & 1.10$^{*}$ (1.07--1.14) & 1.11$^{*}$ (1.06--1.16) \\
\textit{purp}  & 0.84 (0.67--1.06)       & 0.80 (0.63--1.01) \\
rect  & 1.10$^{*}$ (1.06--1.14) & 1.15$^{*}$ (1.09--1.21) \\
forg  & 1.12$^{*}$ (1.08--1.16) & 1.11$^{*}$ (1.05--1.18) \\
port  & 1.21$^{*}$ (1.18--1.24) & 1.22$^{*}$ (1.17--1.27) \\
comp  & 1.18$^{*}$ (1.15--1.21) & 1.19$^{*}$ (1.14--1.23) \\
\textit{hum}   & 1.12$^{*}$ (1.08--1.17) & 1.09$^{*}$ (1.04--1.15) \\
\bottomrule
\end{tabular}

\medskip
\footnotesize{
Odds ratios correspond to the interaction terms from logistic difference-in-differences models estimated on the balanced panel ($11{,}800$ websites; $23{,}600$ policy--snapshot observations). Standard errors are clustered at the website (\texttt{origin}) level. \\
$^{*}$ indicates statistical significance after Benjamini--Hochberg~\cite{Benjamini1995FDR} false discovery rate correction at $\alpha=0.05$, applied separately to \texttt{CH} vs.\ \texttt{EU} and \texttt{CH \& EU} vs.\ \texttt{EU}.
}
\caption{Logistic difference-in-differences.}
\label{tab:did-logit-results}
\end{table}

The interaction terms capture differential temporal change relative to the EU baseline on the odds scale. Table~\ref{tab:did-logit-results} reports the resulting odds ratios. The qualitative pattern mirrors the primary results: interaction effects for the \texttt{CH} and \texttt{CH \& EU} groups are positive and statistically significant for all obligations except \texttt{purp}. As in the primary specification, we control the expected false discovery rate using the Benjamini--Hochberg procedure at $\alpha=0.05$, applied separately to the two families of interaction terms. The logistic specification confirms that the differential increases documented in Section~\ref{sec:results} do not depend on the linear probability formulation.

\subsection{Sample Generator Interface (SwissAnwalt)}\label{app:swissanwalt}

\begin{figure}[H]
    \centering
    \includegraphics[width=1\linewidth]{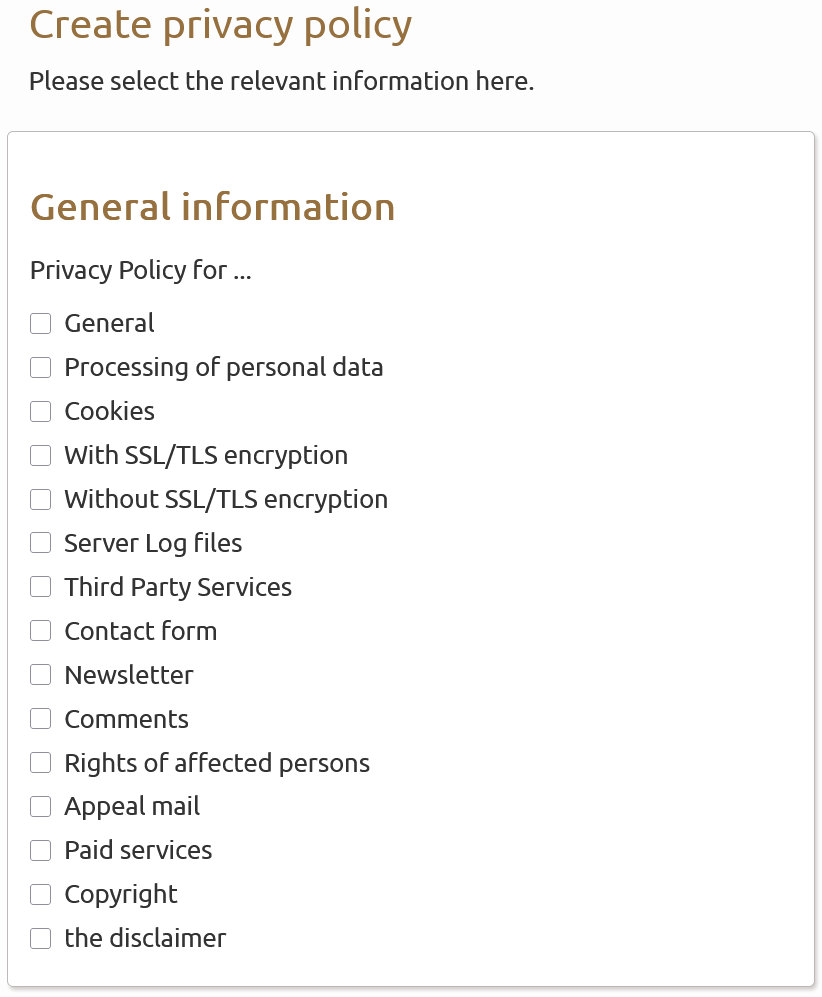}
    \caption{Translated excerpt from the latest available version of the SwissAnwalt policy generator (March 2023).}
    \label{fig:swissanwalt}
    \Description{Screenshot of an excerpt from SwissAnwalt's policy generator from March 2023. The image depicts different boxes (e.g., cookies) that the user can tick to generate a policy.}
\end{figure}

Figure \ref{fig:swissanwalt} displays part of the translated user interface for the most widely used (yet no longer operational) generator in our dataset, ``SwissAnwalt'' (retrieved using the Wayback Machine~\cite{swissanwalt-datenschutz-2023}). To create a policy, a user needed to tick all boxes that applied to their website, e.g., the use of a newsletter, and to provide information about the identity of the controller (not visible in the screenshot). While generating a policy was free, SwissAnwalt's business model seemed to rely on (1) using references in generated policies as advertising and (2) referring users to the network of lawyers by whom it was provided, who offered their services for a fee.

\subsection{Detailed Generator Disclosure Statistics}\label{app:generator-details}

Table~\ref{tab:rank_generator_disclosures} confirms that generator use is associated with increased mean average disclosure rates \textit{independent} from the website popularity rank (CrUX). While generators are most prevalent among less visited websites, these smaller websites also exhibit the highest disclosure increases when using a generator. Moreover, in Section~\ref{sec:generators}, we have provided the disclosure metrics for the five most widely used generators in Table~\ref{tab:top5-generator-obligation-single}. We provide the full analysis for all generators in Table~\ref{tab:generator-obligation-single}.

\begin{table*}[ht]
\centering
\begin{tabular}{lcccccc}
\toprule
 & \multicolumn{2}{c}{\textbf{Top 5k rank}} & \multicolumn{2}{c}{\textbf{Top 10k--50k rank}} & \multicolumn{2}{c}{\textbf{Top 100k+ rank}} \\
\cmidrule(lr){2-3} \cmidrule(lr){4-5} \cmidrule(lr){6-7}
Obligation & No generator & Generator used & No generator & Generator used & No generator & Generator used \\
\midrule
\textit{contr} & 86.3\% & 97.9\% & 79.7\% & 86.8\% & 76.1\% & 89.3\% \\
\textit{purp} & 98.0\% & 100.0\% & 98.7\% & 99.9\% & 98.7\% & 100.0\% \\
rect & 84.9\% & 89.4\% & 80.7\% & 85.1\% & 77.4\% & 82.1\% \\
forg & 86.6\% & 89.4\% & 82.6\% & 85.4\% & 79.9\% & 82.2\% \\
port & 65.6\% & 78.7\% & 54.6\% & 69.8\% & 55.8\% & 67.7\% \\
comp & 66.9\% & 83.0\% & 52.6\% & 69.5\% & 52.2\% & 67.1\% \\
\textit{hum} & 34.6\% & 14.9\% & 22.6\% & 13.9\% & 20.1\% & 16.3\% \\
\midrule
\textbf{Average} & 74.7\% & 79.0\% & 67.3\% & 72.9\% & 65.7\% & 72.1\% \\
\textbf{Total policies} & 1061 & 47 & 6224 & 1000 & 5028 & 973 \\
\bottomrule
\end{tabular}
\medskip
\caption{Disclosure rates per obligation by rank group and generator use (October 2023).}
\label{tab:rank_generator_disclosures}
\end{table*}

\begin{table*}[ht]
\centering
\begin{tabular}{l|rrrrrrr|rrr}
\toprule
Generator & \textit{contr} & \textit{purp} & rect & forg & port & comp & \textit{hum} & \textbf{Average} & \textbf{Policies} & \textbf{Market Share} \\
\midrule
SwissAnwalt (CH) & 92.5\% & 100.0\% & 57.8\% & 58.3\% & 45.5\% & 46.1\% & 1.4\% & 57.4\% &   657 & 32.7\% \\
Datenschutzpartner (CH) & 99.1\% & 99.7\% & 99.7\% & 99.7\% & 89.3\% & 99.7\% & 0.6\% & 84.0\% &   337 & 16.7\% \\
PrivacyBee (CH) & 99.5\% & 100.0\% & 99.5\% & 99.5\% & 98.6\% & 99.5\% & 47.7\% & 92.1\% &   222 & 11.0\% \\
eRecht24 (DE) & 5.8\% & 100.0\% & 98.1\% & 98.1\% & 5.8\% & 5.8\% & 1.3\% & 45.0\% &   156 & 7.8\% \\
DGD (DE) & 98.3\% & 100.0\% & 98.3\% & 98.3\% & 97.5\% & 95.0\% & 99.2\% & 98.1\% &   121 & 6.0\% \\
BrainBox (CH) & 94.2\% & 100.0\% & 76.0\% & 75.2\% & 75.2\% & 73.6\% & 1.7\% & 70.8\% &   121 & 6.0\% \\
activeMind (DE) & 88.8\% & 100.0\% & 98.0\% & 98.0\% & 83.7\% & 75.5\% & 21.4\% & 80.8\% &    98 & 4.9\% \\
Schwenke (DE) & 89.9\% & 100.0\% & 96.2\% & 97.5\% & 89.9\% & 89.9\% & 12.7\% & 82.3\% &    79 & 3.9\% \\
AdSimple (DE) & 92.9\% & 100.0\% & 98.2\% & 100.0\% & 100.0\% & 35.7\% & 25.0\% & 78.8\% &    56 & 2.8\% \\
WeissPartner (DE) & 97.7\% & 100.0\% & 100.0\% & 100.0\% & 86.4\% & 86.4\% & 2.3\% & 81.8\% &    44 & 2.2\% \\
Iubenda (IT) & 100.0\% & 100.0\% & 79.1\% & 79.1\% & 79.1\% & 79.1\% & 14.0\% & 75.7\% &    43 & 2.1\% \\
LegallyOK (CH) & 100.0\% & 100.0\% & 100.0\% & 100.0\% & 89.7\% & 100.0\% & 15.4\% & 86.4\% &    39 & 1.9\% \\
MeinDatenschutz (DE) & 100.0\% & 100.0\% & 100.0\% & 100.0\% & 87.1\% & 77.4\% & 6.5\% & 81.6\% &    31 & 1.5\% \\
TrustedShops (DE) & 50.0\% & 100.0\% & 100.0\% & 100.0\% & 100.0\% & 100.0\% & 25.0\% & 82.1\% &     8 & 0.4\% \\
\bottomrule
\end{tabular}
\medskip

\caption{Disclosures by individual generator and obligation for policies using a single generator (October 2023).}
\label{tab:generator-obligation-single}
\end{table*}

\end{document}